\documentclass{article}

\usepackage[preprint]{preprint}

\usepackage[utf8]{inputenc} 
\usepackage[T1]{fontenc}    
\usepackage{hyperref}       
\usepackage{url}            
\usepackage{booktabs}       
\usepackage{amsfonts}       
\usepackage{nicefrac}       
\usepackage{microtype}      
\usepackage{xcolor}         

\usepackage{microtype}
\usepackage{graphicx}
\usepackage{subfigure}
\usepackage{booktabs} 

\usepackage{hyperref}

\usepackage{mathtools}
\usepackage{amsthm}

\usepackage{times}
\usepackage{soul}
\usepackage{url}
\usepackage[utf8]{inputenc}
\usepackage[small]{caption}
\usepackage{graphicx}
\usepackage{booktabs}
\usepackage{algorithm}
\usepackage{algorithmic}
\usepackage{wrapfig}
\usepackage{subfigure}
\usepackage{lipsum}
\usepackage{amsmath,amsfonts,amssymb}
\DeclareMathOperator*{\argmax}{arg\,max}

\usepackage{multirow}
\usepackage{subfigure}
\usepackage{color}
\usepackage{bbm} 
\makeatletter

\newcommand{\Rmnum}[1]{\expandafter\@slowromancap\romannumeral #1@}
\makeatother

\usepackage[capitalize,noabbrev]{cleveref}

\theoremstyle{plain}

\theoremstyle{definition}

\theoremstyle{remark}

\usepackage[textsize=tiny]{todonotes}

\title{FedNoiL: A Simple Two-Level Sampling Method for Federated Learning with Noisy Labels}

\author{Zhuowei Wang$^{1}$, Tianyi Zhou$^{2,3}$, Guodong Long$^1$, Bo Han$^4$, Jing Jiang$^1$\\
$^1$Australian Artificial Intelligence Institute, University of Technology Sydney\\
$^2$University of Washington, Seattle, $^3$University of Maryland, College Park\\
$^4$Hong Kong Baptist University\\
zhuowei.wang@student.uts.edu.au,
\{guodong.long, jing.jiang\}@uts.edu.au,\\ tianyizh@uw.edu, bhanml@comp.hkbu.edu.hk
}

\begin{document}

\maketitle

\begin{abstract}
 Federated learning (FL) aims at training a global model on the server side while the training data are collected and located at the local devices. Hence, the labels in practice are usually annotated by clients of varying expertise or criteria and thus contain different amounts of noises. Local training on noisy labels can easily result in overfitting to noisy labels, which is devastating to the global model through aggregation. Although recent robust FL methods take malicious clients into account, they have not addressed local noisy labels on each device and the impact on the global model. In this paper, we develop a simple two-level sampling method ``FedNoiL'' that (1) selects clients for more robust global aggregation on the server; and (2) selects clean labels and correct pseudo-labels at the client end for more robust local training. The sampling probabilities are built upon clean label detection by the global model. Moreover, we investigate different schedules changing the local epochs between aggregations over the course of FL, which notably improves the communication and computation efficiency in noisy label setting. In experiments with homogeneous/heterogeneous data distributions and noise ratios, we observed that direct combinations of SOTA FL methods with SOTA noisy-label learning methods can easily fail but our method consistently achieves better and robust performance.\looseness-1
\end{abstract}

\section{Introduction}
\vspace{-0.5em}
Unlike centralized learning that both the model and training data are located at the server side, various practical applications (e.g., for personal devices~\cite{konevcny2016federated}, financial institutions~\cite{long2020federated}, or hospitals~\cite{guo2021multi}) rely on federated learning (FL), where the training data are collected and stay at the local devices for local model training only, while the global model is updated in a data-free manner by aggregating the local models periodically. However, in practical applications, the labels of local data are not annotated by the same client according to the same criterion so they inevitably contain different noises across clients, e.g., their noise ratios are heterogeneous.

It has been widely demonstrated~\cite{zhang2016understanding,arpit2017closer} that the noisy labels can cause catastrophic failure of machine learning models and quickly results in overfitting to wrong labels. In FL setting, this problem can be more devastating because the server cannot directly access data and filter out the noises but relies on the local updates, which is unfortunately more prone to noisy labels without sufficient clean data per local client. 
Although robust FL~\cite{krum,yin2018byzantine,pillutla2019robust} studies how to reduce the impact of malicious clients on the global aggregation, they only apply client-level operations on the server side but ignore the deterioration of noise labels to local model training. Hence, they lack the interaction between global model aggregation and local training for combating label noise. 

A straightforward strategy to address the noisy label challenge in FL might be replacing the local training on each client with an existing noisy-label learning (NLL) algorithm equipped with clean label detection~\cite{han2018co}, transition matrix estimation~\cite{patrini2017making}, or semi-supervised learning~\cite{dividemix}. However, such na\"{\i}ve combination cannot effectively address the challenge because: (1) the data on some local devices are often insufficient and might be highly noisy, in which case the NLL algorithm can easily fail with poor models overfitting to the noises; and (2) the noise ratio can vary drastically across clients (i.e., heterogeneous noises) so those poor local models in (1) are poisoning to the global model that will be broadcast to contaminate other higher-quality local models. Moreover, it is challenging to apply NLL methods globally and they are not designed for non-IID data distributions over heterogeneous clients~\cite{kairouz2019advances}. 

In this paper, we propose a \textbf{simple but effective} method for federated learning with noisy labels (FedNoiL) to address the above challenges. FedNoiL applies a two-level sampling in the existing FL framework: (1) to improve the robustness of global aggregation, it selects higher-quality local models by sampling clients with fewer noise labels; and (2) to rule out noisy labels from the local update, it selects clean labels by local data sampling. The sampling probabilities for both client and local-data sampling are computed from the confidence of the global model, which is less prone to the noise labels than local models and reflects the correctness of labels. 
In addition, we boost the performance by semi-supervised learning that treats the selected samples as labeled data and the rest as unlabeled data with pseudo-labels produced by the global model.

Since noisy label learning usually converges much slower and is less stable than learning with clean labels~\cite{nguyen2019self}, another key challenge for FedNoiL is to improve the computational and communication efficiency, which is essential to FL applications on edge devices. In this paper, we study the schedules of local epochs between two global aggregations and discover that \textbf{starting from more local epochs and progressively reducing the number by a decaying function (e.g., logarithm or cosine) achieve better trade-off between computation and communication efficiency}  than fixing the number of local epochs. Specifically, sufficient local epochs between earlier aggregation rounds are necessary for the convergence of local NLL. This impacts the quality of the aggregated global model and the estimation of sampling probabilities in FedNoiL. Thereby, as the quality of selected clients and samples improves, we can reduce the local epochs to encourage more frequent knowledge sharing across clients and boost the global model performance.

In experiments, we evaluate FedNoiL and compare it with various baselines that combine the SoTA FL and NLL approaches in both IID and non-IID settings of several FL benchmarks, which are modified by adding different types and proportions of label noises to the local data of different devices. Extensive experiments demonstrate that \textbf{trivial combination of existing FL and NLL methods is unstable and can easily fail, while FedNoiL consistently outperforms them by a large margin in all cases}. Moreover, we conduct thorough ablation studies and fine-grained empirical analyses of each key component in FedNoiL, which unveil their indispensable roles in combating the noisy labels in FL.\looseness-1  

\vspace{-0.5em}
\section{Related work}
\vspace{-0.5em}
\paragraph{Federated Learning}
FL~\cite{fedavg,yang2019federated,li2020federated} enables multiple clients to collaborate on distributed learning of a global model without sharing local data. 
In FL, the techniques in personalized FL and robust FL are related to noisy labels. 
Robust FL aims at improving the robustness of FL to malicious clients, e.g., robust aggregation by coordinate-wise median or coordinate-wise trimmed mean of local updates~\cite{yin2018byzantine}, clustering clients into a benign group and a malicious group~\cite{krum,cfl}, etc. 
They mainly rely on client-level operations on the server side, while we additionally study the impact of noisy labels via sample-level operations on each local client.
Personalized FL aims at learning a personalized model for each client in heterogeneous FL where clients differ on data distributions but can still share knowledge through communication with the server.
It has been studied through interpolation between global and local models~\cite{apfl}, meta-learning~\cite{khodak2019adaptive,fallah2020personalized}, multi-task learning~\cite{smith2017federated,dinh2020personalized}, representation learning~\cite{arivazhagan2019federated,collins2021exploiting}, etc. They are related to our problem in addressing the statistical heterogeneity challenge in FL. 
Instead, we study a more challenging and practical setting with noisy labels in the heterogeneous setting. Moreover, we consider heterogeneous distribution of noises across clients.

\textbf{Noisy Label Learning (NLL)}
NLL in centralized learning has been broadly studied recently~\cite{patrini2017making,ren2018learning,zhang2018generalized,tanaka2018joint,wang2019symmetric}.  
One widely applied strategy is clean label selection. {\em Mentornet}~\cite{jiang2018mentornet} aims to downweigh samples with incorrect labels. {\em Co-teaching}~\cite{han2018co} lets two networks to select clean training data for each other. {\em JoCoR}~\cite{wei2020combating} selects small-loss (cross-entropy and co-regularization loss) samples as clean data, while {\em JoSRC}~\cite{josrc} selects data based on  Jensen-Shannon divergence. 
Another line of works~\cite{dividemix,nguyen2019self} applies semi-supervised learning~\cite{berthelot2019mixmatch,fixmatch} that treats wrongly-labeled samples as unlabeled and assigns them with pseudo labels. 
{\em RoCL}~\cite{zhou2020robust} develops a curriculum from the former (supervised learning on clean data) to the latter (pseudo-label learning on noisy data). 
However, existing NLL methods are not developed for FL and require much more data than those on each client in FL.\looseness-1

\textbf{Label Noise in Federated Learning}
There have been only a few attempts on FL with noisy labels. 
\cite{yang2020robust} proposes to interchange both class centroids of local data and local models, which may put client privacy at risk since raw images can be recovered from the centroids~\cite{zhao2020makes}. Moreover, it assumes that all clients share the same noise ratio, which is less practical. 
\cite{chen2020focus,tuor2021overcoming} require a public clean dataset on the server.
\cite{li2021federated} introduces a robust aggregation method to select clients with less noisy labels but does not have sample-level operations (e.g., clean data detection) per client.
To our best knowledge, our method is the first to integrate \textbf{client selection} and \textbf{client-side data selection} to address FL with noisy labels in the vanilla FL setting \textbf{without extra clean data or sending class centroids}.

\section{Problem Formulation}

We consider image classification of $C$ classes in FL, where a global model $\theta$ is periodically updated by aggregating the $K$ local models $\{\theta_k\}^{K}_{k=1}$ trained on $K$ clients. 
Let $D_k=\{(x_{k,i}, \tilde{y}_{k,i})\}^{n_{k}}_{i=1}$ be the local dataset containing $n_k$ samples on client-$k$, where $x_{k,i}$ is the $i$-th training sample and $\tilde{y}_{k,i}$ is the given label that can be the same as the correct label $y_{k,i}$ or a noisy one $\tilde{y}_{k,i}\neq y_{k,i}$.
In FL, the global model $\theta$ aims at minimizing the total loss over all clients, i.e.,
\begin{equation}
\label{eq:global}
\min_{\theta} \mathcal{L}(\theta)\triangleq\sum_{k=1}^{K} \frac{n_{k}}{N} \mathcal{L}_{k}(\theta),~N=\sum_{k=1}^{K} n_{k},
\end{equation}
where $\mathcal{L}_{k}(\theta)$ is the local loss on $D_k$ for client-$k$, i.e.,
\begin{equation}
\label{eq:local}
 \mathcal{L}_{k}(\theta)=\frac{1}{n_{k}} \sum_{i=1}^{n_{k}} \mathcal{L}_{CE}\left(\tilde{y}_{k,i}, F(x_{k,i};\theta)\right),
\end{equation}
where $\mathcal{L}_{CE}$ is the cross entropy loss and $F(\cdot;\theta)$ is the model predicting class probabilities.
Most FL methods address the above optimization problem by iterating between local update per client (initialized by the global model) and global aggregation of uploaded local models. 
Specifically, the local update on device-$k$ starts from initializing $\theta_k\leftarrow \theta$ and then minimizes the local loss $\mathcal{L}_{k}$ by multiple epochs of (stochastic) gradient descent on $D_k$ with each step as
\begin{equation}
\label{eq:local_update}
\theta_{k} \leftarrow \theta_{k}-\eta \nabla_\theta \mathcal{L}_{k}(\theta_k),
\end{equation}
where $\eta$ is the local learning rate. The central server then aggregates the local models by:
\begin{equation}
\label{eq:aggregation}
\theta\leftarrow \sum_{k=1}^{K} \frac{n_{k}}{N} \theta_{k}.
\end{equation}
The server then broadcasts $\theta$ to all clients as the initialization for the next round of local update. 
However, in the presence of noisy labels $\tilde{y}_{k,i}\neq y_{k,i}$, 
the local update can quickly overfit to the label noise in the local dataset $D_k$~\cite{zhang2016understanding}, resulting in some poor local models. Moreover, directly aggregating these models in Eq.~\eqref{eq:aggregation} leads to a global model with poor generalization performance. Therefore, an intuitive motivation to alleviate the above problem is to select high-quality local models only for aggregation and select clean labels on each device to update the local model.\looseness-1


\vspace{-0.5em}
\section{Proposed Method}
\vspace{-0.5em}
In this section, we introduce ``FedNoiL'', a simple but effective federated NLL method based on a novel two-level sampling strategy, which first conducts client sampling to select higher-quality local models for robust global aggregation, followed by local-data sampling that selects clean samples and correct pseudo-labels of local update on each client. Moreover, we explore different schedules of varying local epochs between aggregations over the course of FL, which effectively reduce the communication and computation cost of federated NLL. 

\vspace{-0.5em}
\subsection{Client Sampling}
\vspace{-0.5em}
\label{inter-client}

\begin{wrapfigure}{R}{0.6\textwidth}
\vspace{-18pt}
\begin{minipage}{0.6\textwidth}

    \begin{algorithm}[H]
    \caption{Federated Learning with Noisy Labels}
    \label{alg:seminll}
    
    \begin{algorithmic}[1]
        \STATE {\bf Input:} $K$ clients with local datasets $\{D_k\}_{k=1}^K$, communication rounds $R$, a schedule of local epochs.
        \STATE {\bf Initialize:} global model $\theta$.
        \FOR{$r=0, 1,\dots,R$}
            \FOR{Client $k \in [K]~\textbf{\mbox{in parallel}}$}
                \STATE Calculate $p_k$ using Eq.~\eqref{eqn:first_layer} 
            \ENDFOR
            \STATE Server samples clients $S\subseteq [K]$ according to $p_k$
            \STATE Schedule determines the number of local epochs $T_r$
            \FOR{Client $k \in S~\textbf{\mbox{in parallel}}$}
                \STATE Initialize local model $\theta_k\leftarrow \theta$. 
                \FOR{$t=1,\dots,T_r$}
                    \STATE Calculate $p_{k,i}$ using Eq.~\eqref{eqn:second_layer} 
                    \STATE Client-$k$ samples labeled data  $D_k^\mathrm{x}\subseteq D_k$ by $p_{k,i}$
                    \STATE Client-$k$ selects a subset of $D_k^\mathrm{u}$ for pseudo-labeling and semi-supervised learning
                    \STATE Update $\theta_k$ using Eq.~\eqref{eq:local_update} 
                \ENDFOR
            \ENDFOR
            \STATE Server updates $\theta$ by global aggregation in Eq.~\eqref{eq:global}
        \ENDFOR
    \end{algorithmic}
    \end{algorithm}
  
\end{minipage}
\vspace{-18pt}
\end{wrapfigure}

We consider a challenging setting that allows different noise ratios across clients. A client with lower noise ratio usually learns a better model faster than the ones with higher noise ratios. Therefore, the global model aggregation produces a more robust and better global model if aggregating higher-quality local models associated with lower noise ratios. A better global model is essentially helpful to combat the noisy labels in local update. Hence, we need to first estimate the noise ratios of each client.


We estimate the noise ratio of a client-$k$ by evaluating the correctness of each sample $x_{k,i}$'s label $\tilde{y}_{k,i}$ using the confidence computed on the latest global model, i.e., $F(x_{k,i};\theta)[\tilde{y}_{k,i}]$. 
Since the global model at the early stage is underconfident, we can apply a low temperature $\tau$ in softmax to better distinguish clean and noisy samples. The confidence score is consistent with the small-loss criterion widely used in NLL. It is also pointed out by~\cite{bai2021understanding} that the samples with high confidence have higher probabilities of being clean. We choose to compute it on the global model since it is more robust than local models, which are more prone to possibly high noises. 
Given the confidence scores for all samples in $D_k$, we can estimate the number of clean labels on client-$k$ as $s^k$ by summing up the confidence scores of all the samples on the client.
Each client can compute the score $s_k$ using its received global model and send $s_k$ back to the server. On the server side, we then compute a probability $p_k$ for each client-$k$ to be selected for aggregation,
\begin{equation}
\label{eqn:first_layer}
p_k = \frac {\sum_{i=1}^{n_k} F(x_{k,i};\theta)[\tilde{y}_{k,i}]}{\sum_{j=1}^{K} \sum_{l=1}^{n_j} F(x_{j,l};\theta)[\tilde{y}_{j,l}]},~~s_k=\sum_{i=1}^{n_k} F(x_{k,i};\theta)[\tilde{y}_{k,i}].
\end{equation}
Hence, in each global aggregation round, we only sample a subset of clients and aggregate their local models only to obtain the latest global model, which is more robust to higher-noise clients than the global average of all local models. By broadcasting such an accurate global model to all clients, the server will then receive a better estimation of noise ratios for clients in the next round. 

\subsection{Local-data sampling} \label{intra_client}
In order to exclude the noisy labels from the local model update per client and obtain a high-quality local model, we further propose to sample training data for each local client according to their confidence scores computed by the global model, which is usually more accurate than local models. Specifically, similar to the small-loss criterion in NLL, samples with higher confidence score are selected with higher probability. The probability of selecting $(x_{k,i}, \tilde{y}_{k,i})$ for training $\theta_k$ is computed locally on client-$k$ as
\begin{equation}
\label{eqn:second_layer}
p_{k,i} = \frac {F(x_{k,i};\theta)[\tilde{y}_{k,i}]}{\sum_{j=1}^{n_k} F(x_{k,j};\theta)[\tilde{y}_{k,j}]}.
\end{equation}

A bottleneck of directly applying NLL to local update on each device in FL is the insufficient local training data, which can easily lead to the overfitting of local models to noises. For samples with incorrect given labels, their input features $x_{k,i}$ contain even richer information than their labels but cannot be fully leveraged if we only select the samples with clean labels for training. 
To tackle this problem, we apply recent semi-supervised learning (SSL) techniques by regarding data sampled according to Eq.~\eqref{eqn:second_layer} as labeled data ${D_k^\mathrm{x}}\subseteq D_k$ and other noisy samples as unlabeled data ${D_k^\mathrm{u}}=D_k\backslash {D_k^\mathrm{x}}$. In particular, we adopt {\em FixMatch}~\cite{fixmatch} that selects the high-confidence samples in ${D_k^\mathrm{u}}$ and relabel them by the model predictions (i.e., pseudo labels) estimated over multiple weak augmentations $\mathcal{A}_w(\cdot)$. It then minimizes the loss on these ``unlabeled data'' using strong augmentations $\mathcal{A}_s(\cdot)$, i.e.,
\begin{align}
\label{eq:lu}
\mathcal{L}_{\mathrm{u}}(\theta_k)=&\frac{1}{|D_k^\mathrm{u}|}  \sum_{x_{k,i}\in D_k^\mathrm{u}} \mathbbm{1}\left(F(\mathcal{A}_w(x_{k,i});\theta)[\hat y_{k,i}] \geq \xi\right)\cdot\mathcal{L}_{CE}\left(\hat y_{k,i}, F(\mathcal{A}_s(x_{k,i});\theta_k)\right),
\end{align}
where the pseudo label $\hat y_{k,i}$ for $x_{k,i}$ is defined as
\begin{equation}
    \hat y_{k,i}\triangleq \argmax_{j\in\{1,2,\cdots,C\}} F(\mathcal{A}_w(x_{k,i});\theta)[j].
\end{equation}
In addition, for the labeled data ${D_k^\mathrm{x}}$ sampled by the probability in Eq.~\eqref{eqn:second_layer}, we minimize the cross-entropy loss:
\begin{equation}
\label{eq:lx}
\mathcal{L}_{\mathrm{x}}(\theta_k)=\frac{1}{|D_k^\mathrm{x}|}  \sum_{x_{k,i}\in D_k^\mathrm{x}} \mathcal{L}_{CE}\left(\tilde y_{k,i}, F(\mathcal{A}_w(x_{k,i});\theta_k\right).
\end{equation}
The final loss minimized by the local update for client-$k$ combines the above two loss terms, i.e., $\mathcal{L}_{k}(\theta_k) = \mathcal{L}_{\mathrm{x}}(\theta_k) + \lambda_u \mathcal{L}_{\mathrm{u}}(\theta_k)$, where $\lambda_u$ is the weight for unlabeled loss.


\subsection{Schedules of Local Epochs}
\label{local epoch}

\begin{wrapfigure}{r}{0.35\textwidth}
    \vspace{-12pt}
    \includegraphics[width=0.35\textwidth]{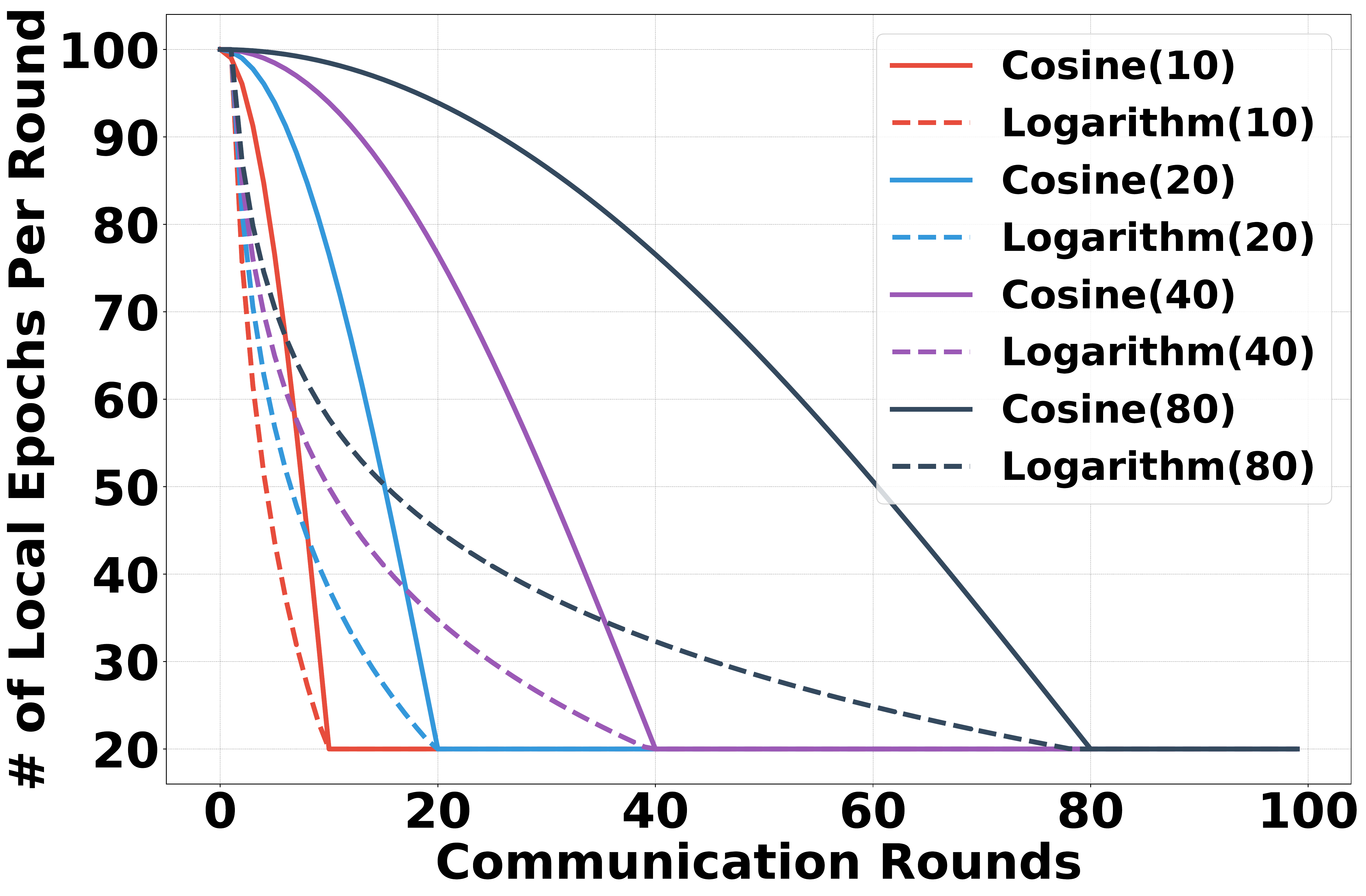}
    \vspace{-15pt}
    \caption{Examples for decaying schedules of local epochs in the form of Cosine($r_{\min}$) and Logarithm($r_{\min}$), which reaches the minimum local epochs $T_{\min}$ at round-$r_{\min}\in\{10,20,40,80\}$.}
    \label{fig:illustration}
    \vspace{-15pt}
\end{wrapfigure} 

Unlike supervised learning with clean labels in previous FL methods, it has been widely observed that NLL and SSL converge much slower and need more epochs during the early stages to train a reliable model used to detect clean labels or produce accurate pseudo labels. In FL, communication cost between server and clients is a major concern of various practical applications. Therefore, in the new FL setting with noisy labels, we propose to start from a large number $T_{\max}$ of local epochs between consecutive rounds of global aggregation in early stages and apply a schedule gradually decreasing the number $T_r$ until reaching a predefined minimum value $T_{\min}$.  
As the quality of sampled clients, selected data, and local models improves, more frequent aggregation and knowledge sharing become helpful to boost the performance of the global model.







We explore two decaying schedules associated with two functions in experiments.
The first schedule is based on cosine function that decreases the local epochs slower during the earlier stages than later, i.e., for round-$r$ of the total $R$ rounds, we apply $T_r$ epochs of local update, i.e.,
\begin{equation}
T_r=\max\left\{T_{\min}+\left(T_{\max}-T_{\min }\right)\cos \left(\frac{r-1}{\psi_1 R} \pi\right), T_{\min}\right\},
\end{equation}
where $\psi_1$ is the coefficient that controls the decay speed of the cosine function.





In contrast, we also consider a decaying schedule based on logarithm function that decreases the local epochs faster during the earlier stages than later, i.e., the number of local epochs at round-$r$ is
\begin{equation}
T_r=\max\left\{T_{\max} - \log_{\psi_2}r, T_{\min}\right\},
\end{equation}
where $\psi_2$ is the coefficient that controls the decay speed of the logarithm function. By choosing different $\psi_1$ and $\psi_2$, we can control the two schedules to reach the minimum epoch number $T_{\min}$ at a specified round $r_{\min}\in \{10, 20, 40, 80\}$. 
Fig.~\ref{fig:illustration} illustrates these examples of cosine and logarithm schedules, which will be evaluated in later experiments and compared to the previously used constant local epochs. 

\section{Experiment}
In this section, we first compare the performance of FedNoiL under different schedules of local epochs. We then compare FedNoiL with the best schedule against direct combinations of SoTA FL and SoTA NLL methods. To demonstrate the practical merits of FedNoiL, we additionally show its effectiveness with more clients, imbalanced samples, and clean local data. Moreover, we present an ablation study of the two-level sampling and the semi-supervised learning.

\vspace{-0.5em}
\subsection{Experimental Setup} \label{experiment setup}
\noindent \textbf{Datasets and Models.} We evaluate our method on three public datasets: FASHION-MNIST~\cite{fmnist}, CIFAR-10~\cite{cifar}, and CIFAR-100~\cite{cifar100}. Two partitioning strategies are used to generate the local datasets: (i) IID: each local client has the same amount of samples in proportion to each class; (ii) Non-IID: we sample $q_j \sim Dir_N(\beta)$ and assign a $q_{j,k}$ proportion of the samples of class $j$ to client $k$, where $Dir(\beta)$ is the Dirichlet distribution with a concentration parameter $\beta$. To add noisy labels, after partitioning, we corrupt these datasets by two widely-used types of noisy labels: symmetric flipping~\cite{rooyen2015learning} and pair flipping~\cite{han2018co}. For the global and local models, we adopt a multi-layer CNN with two convolutional layers followed by one fully-connected layer for FASHION-MNIST, and ResNet18~\cite{he2016deep} for CIFAR-10 and~CIFAR-100.

\noindent \textbf{Heterogeneous noise ratios.} We split all $20$ clients into four groups, and different groups of clients have different noise ratios. According to the overall noise ratio, we evaluate two noise modes and denote them as ``high'' and ``low''. For symmetric flipping noises, the high-noise ratios for the four groups are $0.5, 0.6, 0.7, 0.8$, respectively, while the low-noise ratios are $0.3, 0.4, 0.5, 0.6$. For pair flipping, the high-noise ratios are $0.3, 0.5, 0.6, 0.8$ while the low-noise ratios are $0.3, 0.4, 0.5, 0.6$.

\noindent \textbf{Baselines.} We compare FedNoiL with baselines that \textbf{directly combine existing methods from FL and NLL}, each applying an NLL method to local update within an existing FL algorithm. For FL, we select four algorithms covering representative FL strategies related to FedNoiL: (i) FedAvg~\cite{fedavg}, (ii) personalized FL: APFL~\cite{apfl}, (iii) robust FL: Krum~\cite{krum}, and CFL~\cite{cfl}. For NLL, we consider two SoTA methods: (i) DMix~\cite{dividemix} and (ii) JoSRC~\cite{josrc}. We denote these baselines in the form of ``FL'' $+$ ``NLL''. Moreover, vanilla FedAvg and FedProx are also included as baselines to highlight that the heterogeneous noise ratios create non-trivial challenge to classic FL methods. We also report results on DMix and JoSRC in centralized setting for reference.

\noindent \textbf{Implementation Details.}
We set 20 clients for all datasets and sample $30\%$ of clients per round of FL. We allocate 600, 500, and 1000 samples to each client for FASHION-MNIST, CIFAR-10, and CIFAR-100, respectively. We use SGD optimizer with learning rate of $0.05$, weight decay of $0.0001$, and momentum of $0.5$ in all experiments. The batch size is set to $32$. The number of communication rounds is set to 150 for FASHION-MNIST, and 200 for CIFAR-10/100. The number of local epochs for all baselines is set to 30. For the schedules of FedNoiL, we set $T_{\max} = 100$ and $T_{\min} = 20$. The size of the clean subset for local-data sampling is set to $0.35N$ for the high-noise mode, and $0.55N$ for the low-noise mode. The temperature $\tau$ is set to 0.5.

\begin{figure}[h]
    \small
    \vspace{-0.15in}
    \begin{tabular}{c c c c}
        \hspace{-0.1in} \includegraphics[width=0.25\textwidth]{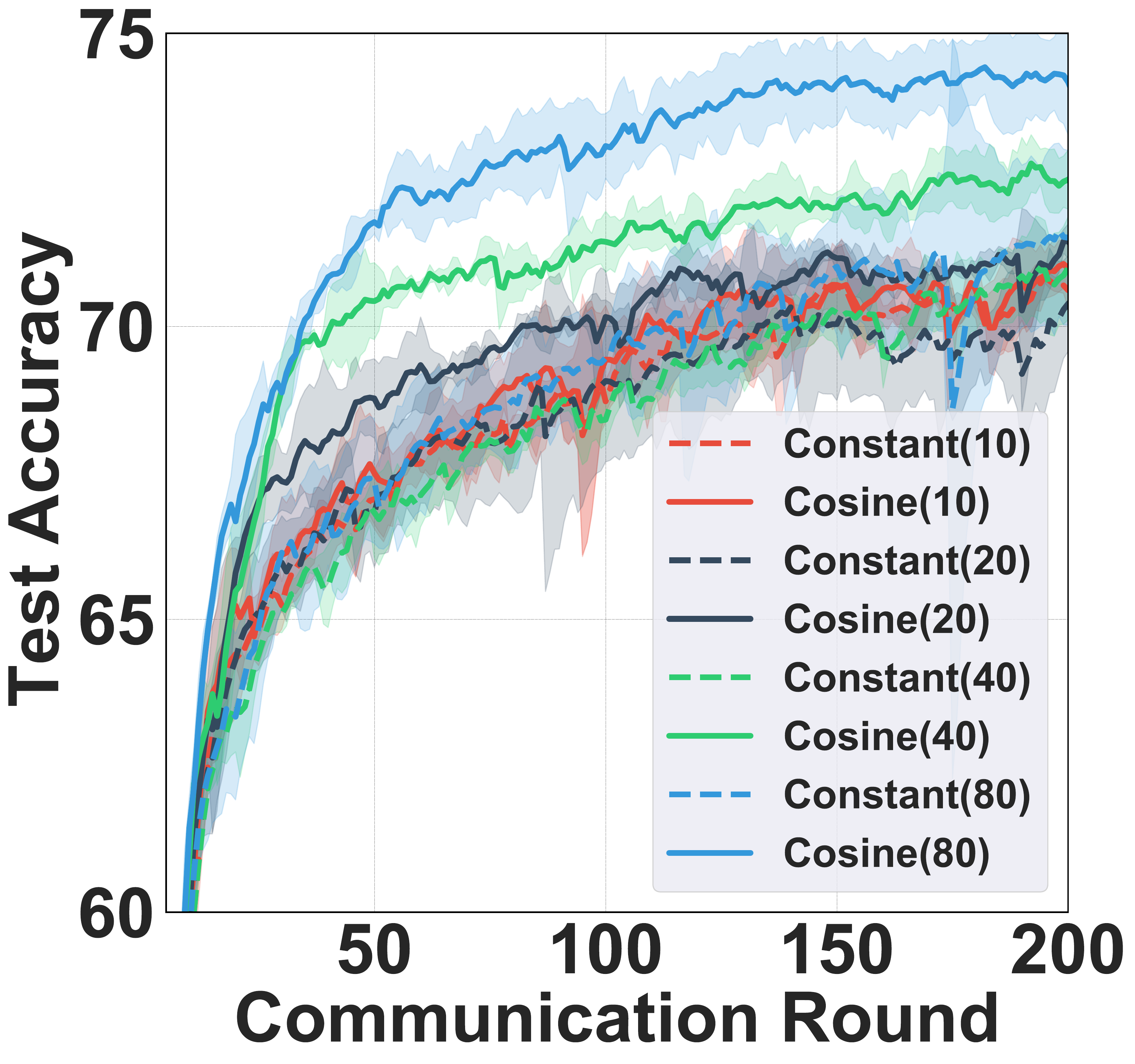}
        & \hspace{-0.18in}\includegraphics[width=0.2365\textwidth]{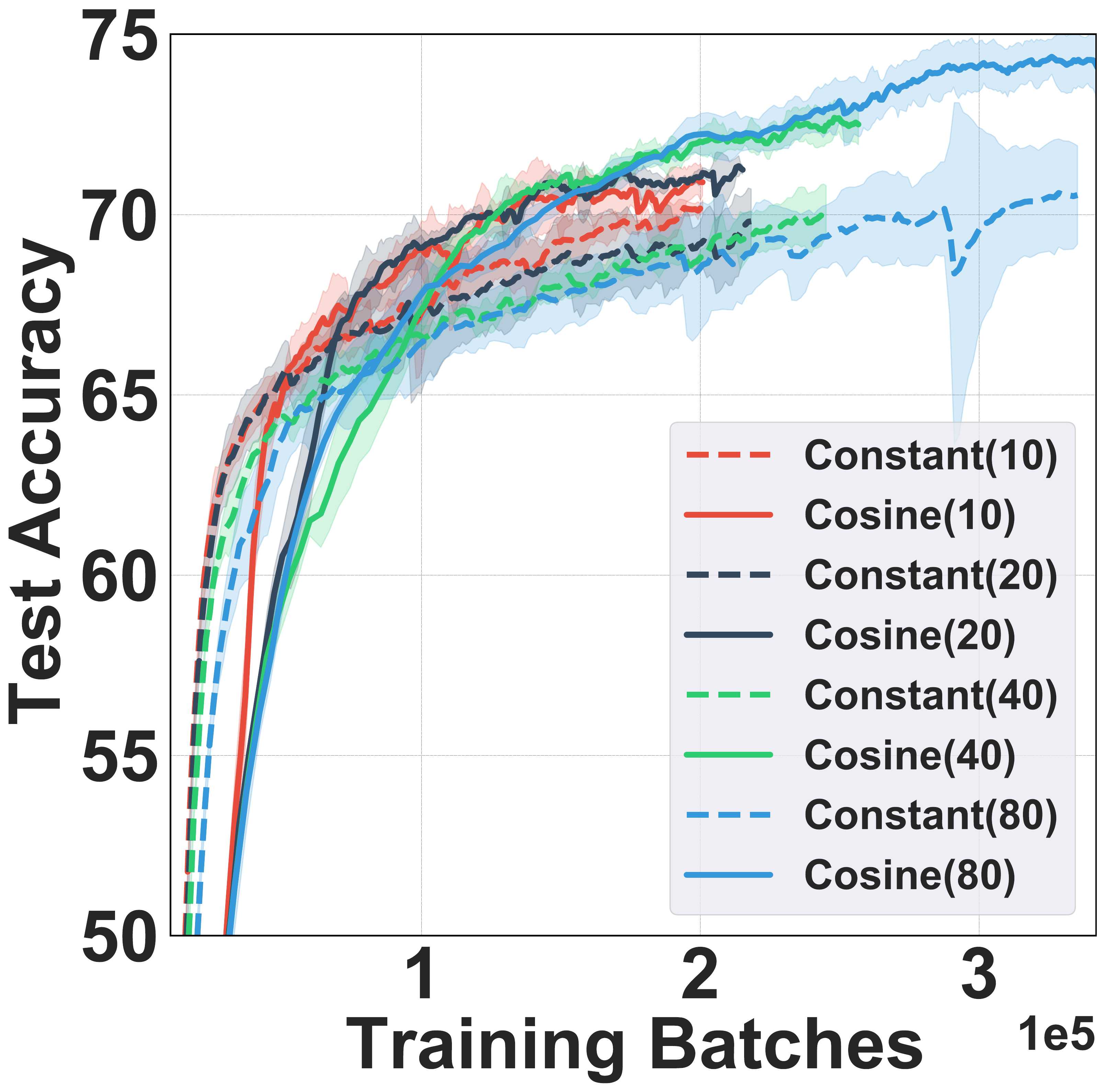}
        & \hspace{0.02in}\includegraphics[width=0.25\textwidth]{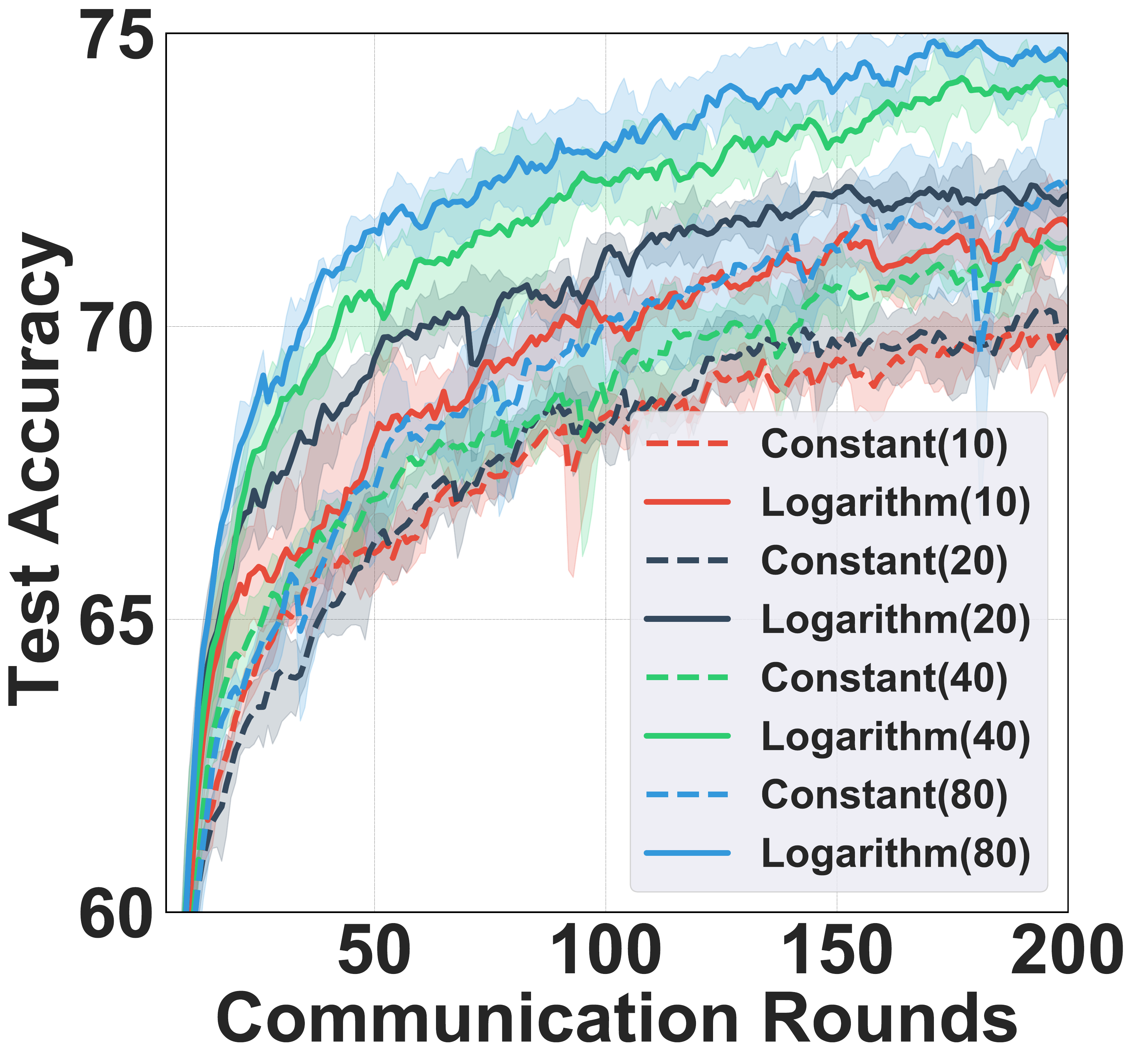}
        & \hspace{-0.18in}\includegraphics[width=0.242\textwidth]{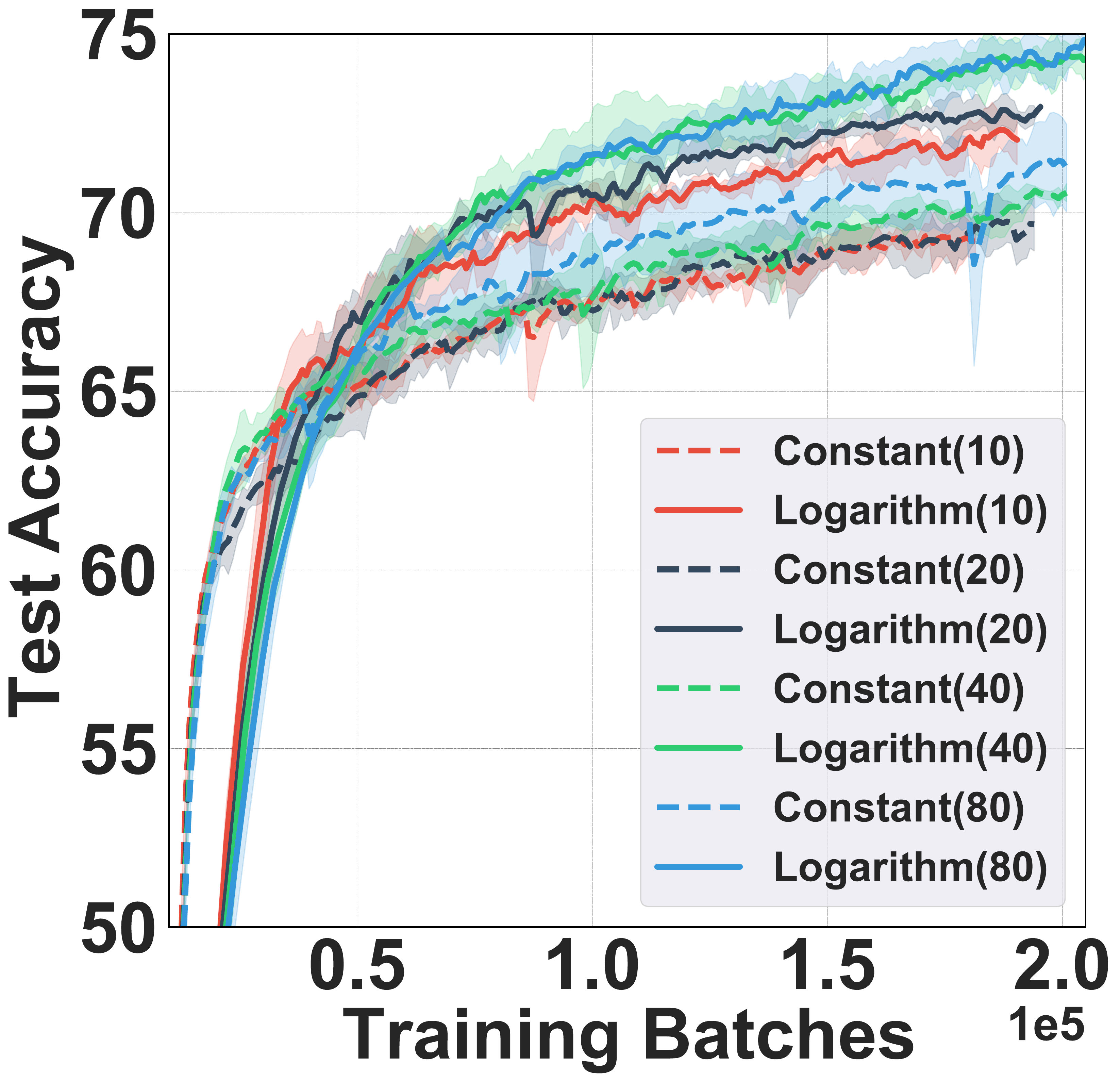}
        \\
        \multicolumn{2}{c}{(a) FedNoiL(\textbf{Cosine})} 
        & \multicolumn{2}{c}{(b) FedNoiL(\textbf{Logarithm})} 
    \end{tabular}
    \vspace{-0.1in}
    \caption{\textbf{Communication (Rounds) and computation (right) efficiency} of FedNoiL using the two proposed decaying schedules of local epochs with different $r_{\min}$ shown in Fig.~\ref{fig:illustration}, compared to constant local epochs matching the total training batches of Cosine($r_{\min}$), on CIFAR-10 with high-noise ratio (symmetric) setting. Colors represent different $r_{\min}$. Cosine and constant schedules are denoted by solid lines and dash lines, respectively.}
    \label{fig:equal}
    \vspace{-0.23in}
\end{figure}

\begin{figure*}[t]
    \vspace{-0.3in}
        \subfigure[Cosine, Logarithm, and Constant\label{fig:parallel_1}]{\includegraphics[width=0.33\textwidth]{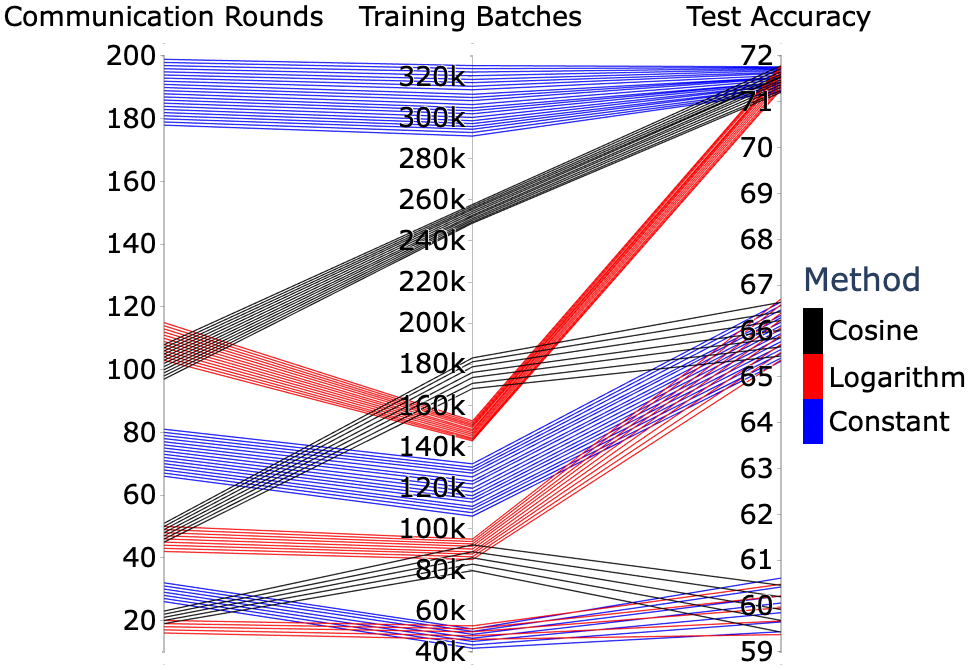}}
        \subfigure[Logarithm $\{20, 40, 80\}$\label{fig:parallel_2}]{\includegraphics[width=0.33\textwidth]{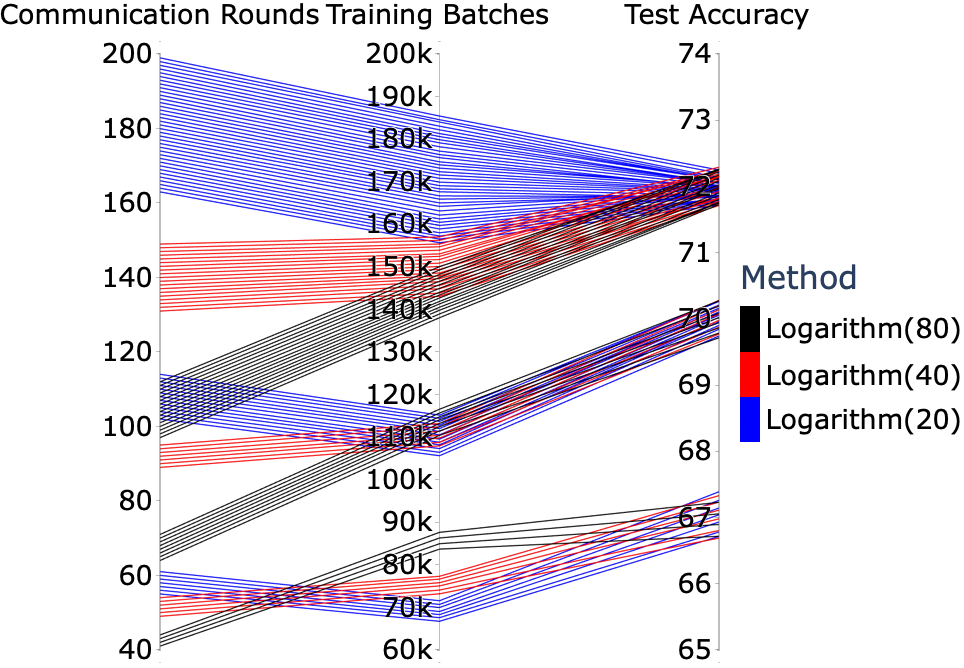}}
        \subfigure[Cosine $\{20, 40, 80\}$\label{fig:parallel_3}]{\includegraphics[width=0.33\textwidth]{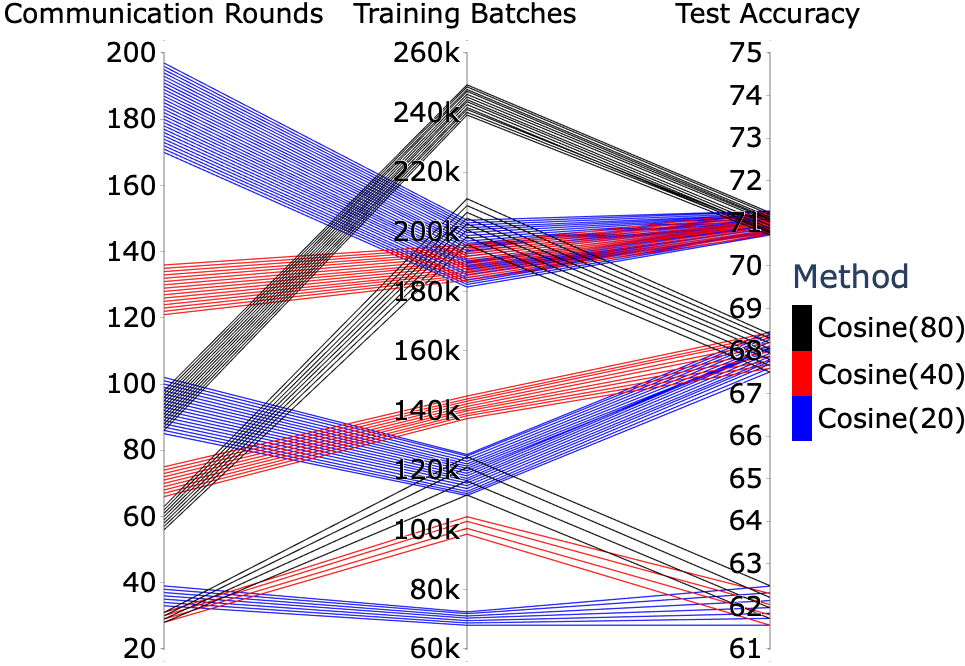}}
    \vspace{-0.15in}
    \caption{Comparison of FedNoiL using different local-epoch schedules on communication, computation, test accuracy, and their trade-off. Each line is a triple recorded at each round. The same color highlights the triples (rounds) from the same scheduling method.}
\end{figure*}

\subsection{Choices of Local Epoch Schedules}
\label{choice_schedule}
We give a brief conclusion on the choices of the local epoch schedules for FedNoiL and leave a detailed analysis to Appendix. As described in Section ~\ref{local epoch}, there are two functions, Cosine and Logarithm, each with four $r_{\min}\in \{10, 20, 40, 80\}$. Moreover, constant schedules refer to FedNoiL with fixed local epochs at each round. They are denoted as Constant($r_{\min}$) and match the total training batches of Cosine($r_{\min}$) or Logarithm($r_{\min}$). (1) The solid lines in Fig.~\ref{fig:equal}(a) represent Cosine schedules with $r_{\min}\in \{10, 20, 40, 80\}$. \textbf{When $r_{\min}$ is set to 80, the Cosine schedule obtains the highest communication and computation efficiency}. (2) The solid lines in Fig.~\ref{fig:equal} (b) represent Logarithm function with $r_{\min}\in \{10, 20, 40, 80\}$. \textbf{When $r_{\min}=80$, the Logarithm schedule also obtains the highest communication and computation-efficiency}. 

\begin{figure}[!h]
    \begin{minipage}[t]{.99\linewidth}
    \vspace{-0.4in}

    \centering
    \begin{table}[H]
        \caption{\textbf{Test accuracy} (\%) of FedNoiL and baselines (direct combination of SOTA FL + SOTA NLL methods, centralized NLL methods, and classic FL methods) under \textbf{IID} and \textbf{Non-IID} settings on three datasets. The final accuracy (mean$\pm$std over five trials) of converged algorithms are reported. \textbf{Methods that do not converge are marked by $\ast$} and the maximum accuracy over all rounds is reported instead.}
        \label{table:iid_main} 
        
        \resizebox{\textwidth}{!}{
        \setlength{\tabcolsep}{0.5mm}{
        \begin{tabular}{l|l|cc|cc|cc|cc}
        \toprule
        \multirow{3}{*}{Datasets} & \multirow{3}{*}{Method} & \multicolumn{4}{c|}{IID} & \multicolumn{4}{c}{Non-IID}  \\ \cline{3-6} \cline{7-10}
        
         &  & \multicolumn{2}{c|}{Symmetric} & \multicolumn{2}{c|}{Pair} & \multicolumn{2}{c|}{Symmetric} & \multicolumn{2}{c}{Pair} \\ \cline{3-6} \cline{7-10}
        & & high & low & high & low & high & low & high & low  \\  \midrule
        
        \multirow{14}{*}{\begin{tabular}[c]{@{}c@{}}FASHION\\MNIST\\ \end{tabular}} 
        & DMix (centralized) & $ 86.67 \pm 0.45 $&$ 88.37 \pm 0.09 $ & $ 85.39 \pm 0.38 $ & $ 86.91 \pm 0.30 $ &$ 85.13 \pm 0.23 $ & $ 87.98 \pm 0.18 $& $ 84.60 \pm 0.39 $ & $ 87.87 \pm 0.19 $\\
        & JoSRC (centralized) &$ 80.66 \pm 0.29 $ & $ 86.29 \pm 0.07 $& $ 83.11 \pm 0.82 $& $ 85.01 \pm 0.81 $ & $ 80.27 \pm 0.92 $& $ 83.62 \pm 0.29 $& $ 78.64\pm 0.83 $& $ 81.03 \pm 0.29 $\\
        \cmidrule{2-10}
        & Vanilla FedAvg & $ 58.30^\ast $ & $ 76.35 ^\ast $ & $ 52.22^\ast  $ & $69.77^\ast  $ &$56.07^\ast  $ & $72.90^\ast  $& $50.67^\ast  $ &$60.79^\ast  $ \\
        & Vanilla FedProx & $ 60.69^\ast $ & $ 70.38^\ast $& $  47.73^\ast $& $   63.66^\ast $ & $   54.40^\ast
        $& $  70.01^\ast $ & $  52.30^\ast $& $  58.09^\ast $\\

        & FedAvg $+$ DMix& $ 70.70 \pm 0.48 $ & $ 78.79 \pm 1.69 $ & $  66.25 \pm 1.97 $ & $ 72.10 \pm 0.49 $ &  $ 67.43 \pm 1.79 $ & $  75.75 \pm 0.70$ & $  59.94 \pm 0.71 $ & $ 65.96 \pm 1.73 $ \\
        & FedAvg $+$ JoSRC & $ 75.54^\ast $ & $  77.54 \pm 1.63 $ & $  75.51^\ast $ & $ 75.68^\ast $  &  $ 72.43^\ast $ & $74.71^\ast  $ & $  74.62^\ast $ & $ 75.48^\ast $\\
        & APFL $+$ DMix & $ 73.32 \pm 1.64 $ & $ 79.07 \pm 1.29  $ & $ 71.46 \pm 1.20  $ & $ 74.69 \pm 1.25 $ &  $ 69.59 \pm 0.91 $ & $ 74.21 \pm 1.45 $ & $ 56.99 \pm 1.27$ & $ 76.63 \pm 2.16 $ \\
        & APFL $+$ JoSRC&   $ 69.63^\ast $ & $ 81.64 \pm 1.26 $ & $ 43.54 \pm 2.42  $ & $ 67.79 \pm 1.87 $ &  $ 66.90 \pm 1.52 $ & $ 78.21 \pm 1.11 $ & $  74.02^\ast $ & $74.60^\ast  $  \\
        & Krum $+$ DMix &  $ 72.18 \pm 0.60 $ & $ 78.69 \pm 1.37 $ & $ 71.54 \pm 1.16  $ & $75.76 \pm 1.61 $ &  $70.36 \pm 0.82  $ & $ 75.52 \pm 0.67 $ & $  42.42 \pm 0.99 $ & $  66.74 \pm 1.53$ \\
        & Krum $+$ JoSRC &  $ 73.33^\ast $ & $ 80.36^\ast $ & $  74.84^\ast $ & $ 75.15^\ast $ &  $ 70.41^\ast $ & $ 73.92^\ast $ & $  70.98^\ast $ & $  75.84^\ast$ \\
        & CFL $+$ DMix & $ 72.27 \pm 0.64 $ & $  78.04 \pm 1.24$ & $  70.68 \pm 1.77 $ & $ 74.68 \pm 2.02 $ &  $ 71.04 \pm 0.52 $ & $ 76.07 \pm 0.59$ & $ 55.51 \pm 0.32  $ & $ 78.75 \pm 1.92 $ \\
        & CFL $+$ JoSRC & $ 75.64^\ast $ & $ 77.71 \pm 1.62 $ & $  74.92^\ast $ & $ 75.33^\ast$ &  $  73.58^\ast$ & $  76.85 \pm 1.42 $ & $  75.72^\ast $ & $76.80^\ast  $ \\
        \cmidrule{2-10}
        
        & \textbf{FedNoiL(Cosine)}&  $ \boldsymbol{83.72 \pm} 0.30 $ & $ 85.62 \pm 0.24 $ & $ 83.88 \pm 0.68  $ & $ 84.73 \pm 0.53 $ &  $\boldsymbol{81.50 \pm 0.34}  $ & $  85.02 \pm 0.63$ & $ 81.95 \pm 1.08  $ & $  84.36 \pm 0.29$ \\
        & \textbf{FedNoiL(Logarithm)}&  $ 82.46 \pm 0.54 $ & $ \boldsymbol{85.91 \pm 0.30} $ & $ \boldsymbol{84.04 \pm 0.45}  $ & $ \boldsymbol{85.22 \pm 0.50} $ &  $80.60 \pm 0.68 $ & $  \boldsymbol{85.38 \pm 0.29}$ & $  \boldsymbol{82.02 \pm 0.66} $ & $ \boldsymbol{84.75 \pm 0.42} $ \\
        \midrule
        
        \multirow{14}{*}{CIFAR-10}
        & DMix (centralized) &$  76.24 \pm 0.52 $ &$  82.14 \pm 0.35 $ & $ 76.19 \pm 0.74 $& $ 80.51 \pm 0.69 $ &$  71.10 \pm 1.09 $ & $  79.60 \pm 0.26 $ &$  73.64 \pm 0.21 $ &$  78.01 \pm 0.34 $ \\
        & JoSRC (centralized) & $ 72.66 \pm 0.76 $ & $ 79.10 \pm 0.98 $& $ 70.71 \pm 1.10 $& $ 75.55 \pm 0.58 $ & $  65.13 \pm 1.21 $& $  75.05 \pm 0.38 $&$  64.01 \pm 0.97 $ & $  70.87 \pm 0.66 $\\
        \cmidrule{2-10}
        & Vanilla FedAvg & $ 42.17 ^\ast $ & $ 55.64^\ast  $ & $ 37.46^\ast  $ & $ 47.57^\ast $ & $ 39.03^\ast $ & $ 50.74^\ast $ & $ 27.90^\ast  $ & $ 55.61^\ast $ \\
        & Vanilla FedProx & $  52.20^\ast $ & $ 54.81^\ast$ & $ 26.43^\ast$ & $ 45.42^\ast$ & $43.98^\ast$ & $56.91^\ast$& $26.17^\ast$ &$54.37^\ast$ \\
        & FedAvg $+$ DMix&  $ 60.97 \pm 1.38 $ & $ 65.85 \pm 1.47 $ & $  48.08^\ast $ & $  62.85 \pm 1.31 $ & $ 58.56 \pm 2.07 $ & $ 64.01 \pm 0.94 $ & $ 24.38 \pm 2.55  $ & $ 55.52 \pm 0.94 $  \\
        & FedAvg $+$ JoSRC &  $ 56.53^\ast $ & $  58.57 \pm 1.43 $ & $  42.84^\ast $ & $ 52.61^\ast $ &  $56.17^\ast  $ & $  59.19^\ast$ & $ 29.45^\ast  $ & $ 50.63^\ast $  \\
        & APFL $+$ DMix &  $ 66.43 \pm 1.81 $ & $ 70.99 \pm 1.61 $ & $  48.38 \pm 1.85 $ & $ 67.58 \pm 1.88 $ &  $60.04 \pm 1.17  $ & $ 67.11 \pm 1.05 $ & $  27.15 \pm 1.26 $ & $ 57.51 \pm 2.33 $ \\
        & APFL $+$ JoSRC  &  $ 33.30 \pm 1.74 $ & $ 50.46 \pm 2.16 $ & $  45.28^\ast $ & $ 53.65^\ast $ & $  29.67 \pm 1.53$ & $ 44.94 \pm 1.68 $ & $  51.62^\ast $ & $ 52.07^\ast$ \\
        & Krum $+$ DMix &  $ 60.66 \pm 1.03 $ & $ 63.05 \pm 1.83 $ & $ 41.75 \pm 1.84 $ & $ 55.00 \pm 1.53 $  & $56.94 \pm 1.61  $ & $ 63.59 \pm 1.48 $  & $19.39 \pm 2.02  $ & $63.35 \pm 2.29  $ \\
        & Krum $+$ JoSRC &  $ 51.63^\ast $ & $ 62.28^\ast $ & $  38.49^\ast $ & $50.63^\ast  $ &  $33.37 \pm 1.65$ & $ 52.97 \pm 2.23 $ & $ 43.83^\ast  $ & $ 46.71^\ast $ \\
        & CFL $+$ DMix &  $ 59.86 \pm 1.05 $ & $ 65.74 \pm 1.51 $ & $  48.80 \pm 1.87 $ & $ 55.60 \pm 1.35 $ &  $ 57.13 \pm 1.39 $ & $ 67.25 \pm 1.54 $ & $  24.62 \pm 1.86 $ & $  56.04 \pm 1.03  $ \\
        & CFL $+$ JoSRC &  $56.55^\ast  $ & $ 58.68 \pm 1.41 $ & $ 53.61^\ast  $ & $ 56.39^\ast $ &  $ 35.39 \pm 2.15 $ & $ 55.00 \pm 1.49 $ & $  52.47^\ast $ & $ 52.51^\ast $ \\
        \cmidrule{2-10}
        
        & \textbf{FedNoiL(Cosine)}&  $ 74.07 \pm 0.49 $ & $ 80.07 \pm 0.51 $ & $  \boldsymbol{73.47 \pm 1.34}$ & $\boldsymbol{77.91 \pm 0.49} $ &  $ 67.49 \pm 1.68 $ & $ 76.21 \pm 0.72 $ & $ \boldsymbol{70.03 \pm 1.13}  $ & $ \boldsymbol{74.19 \pm 0.62} $ \\
        & \textbf{FedNoiL(Logarithm)}&  $ \boldsymbol{74.62 \pm 0.69} $ & $\boldsymbol{80.39 \pm 0.67}  $ & $  71.44 \pm 0.69 $ & $ 75.66 \pm 0.76 $ &  $ \boldsymbol{68.67 \pm 0.87} $ & $ \boldsymbol{76.23 \pm 0.86} $ & $ 68.47 \pm 0.46  $ & $ 72.07 \pm 1.14$ \\
        \midrule
        
        \multirow{14}{*}{CIFAR-100}
        & DMix (centralized) & $44.90 \pm 0.61$& $49.33 \pm 0.43$ & $ 46.04 \pm 0.16 $ & $ 48.78 \pm 1.41 $ &$40.64 \pm 1.01$ &$48.42 \pm 0.21$ & $42.91 \pm 0.29$ &$ 47.55 \pm 0.44$ \\
        & JoSRC (centralized) &$41.12 \pm 0.67$ & $45.93 \pm 0.39$& $ 40.19 \pm 0.69 $ & $ 42.01 \pm 0.67 $ &$35.10 \pm 0.92$ & $42.07 \pm 1.34$& $36.92 \pm 0.41 $&$ 41.00 \pm 0.25$ \\
        \cmidrule{2-10}
        & Vanilla FedAvg & $ 25.06^\ast $ & $ 33.95^\ast $ & $ 20.75^\ast  $ & $ 31.20^\ast $ & $ 24.61^\ast $ & $ 29.90^\ast $ & $  20.43^\ast $ & $ 30.12^\ast $ \\
        & Vanilla FedProx & $28.37^\ast$ & $29.88^\ast$& $ 26.95^\ast$ & $ 28.54^\ast$ & $26.32^\ast $ &$33.07^\ast $  & $ 21.94\ast $ &$29.08^\ast $ \\
        & FedAvg $+$ DMix& $ 39.22 \pm 0.58 $ & $44.74 \pm 0.48  $ & $ 32.10^\ast  $ & $ 39.64^\ast $ & $ 28.61 \pm 0.50 $ & $ 34.16 \pm 0.45 $ & $ 25.39^\ast  $ & $ 32.70^\ast $ \\
        & FedAvg $+$ JoSRC &  $ 26.29 \pm 1.18 $ & $40.35 \pm 0.82  $ & $  22.38^\ast $ & $32.93^\ast  $ & $26.61 \pm 1.15  $ & $ 40.45 \pm 0.68 $ & $ 22.06^\ast  $ & $31.12 \pm 1.80  $ \\
        & APFL $+$ DMix &  $ 34.27 \pm 0.85 $ & $41.97 \pm 1.20  $ & $  33.69^\ast $ & $ 37.27 \pm 1.35$ &  $ 33.37 \pm 1.01 $ & $ 40.02 \pm 0.63 $ & $   24.91^\ast$ & $ 32.21^\ast $ \\
        & APFL $+$ JoSRC&  $ 21.29 \pm 0.87 $ & $ 34.31 \pm 1.31 $ & $ 30.25^\ast  $ & $ 34.07^\ast $ &  $ 20.55 \pm 1.08 $ & $ 33.43 \pm 1.16 $ & $   25.38^\ast$ & $ 29.67^\ast $ \\
        & Krum $+$ DMix &  $ 35.98 \pm 1.31 $ & $ 42.49 \pm 0.50 $ & $ 27.03^\ast  $ & $36.01 \pm 2.70  $ &  $34.01 \pm 1.42  $ & $ 42.09 \pm 0.72 $ & $  23.37^\ast $ & $ 30.28^\ast $ \\
        & Krum $+$ JoSRC &  $ 23.04 \pm 1.60 $ & $ 37.60 \pm 1.29 $ & $ 22.81^\ast $ & $ 32.20^\ast $ &  $23.56 \pm 1.80  $ & $ 37.91 \pm 1.30 $ & $ 21.46^\ast  $ & $  30.62^\ast$ \\
        & CFL $+$ DMix & $ 39.56 \pm 0.53 $ & $ 44.68 \pm 0.57 $ & $  31.63^\ast $ & $ 41.22 \pm 0.65 $ & $ 28.98 \pm 0.42 $ & $43.78 \pm 0.47  $ & $  25.33^\ast $ & $ 33.28 \pm 1.32 $ \\
        & CFL $+$ JoSRC & $26.34 \pm 1.18  $ & $ 40.36 \pm 0.89 $ & $  22.26^\ast  $ & $32.86^\ast  $ & $ 26.55 \pm 1.06 $ & $ 40.32 \pm 0.64 $ & $   21.59^\ast $ & $ 31.89^\ast $ \\
        \cmidrule{2-10}
        
        & \textbf{FedNoiL(Cosine)}&  $\boldsymbol{41.94 \pm 0.80} $ & $ \boldsymbol{47.96 \pm 0.70}  $ & $ \boldsymbol{41.93 \pm 0.77}   $ & $ 44.61 \pm 0.40  $ &  $  \boldsymbol{35.71 \pm 0.71} $ & $ \boldsymbol{46.87 \pm 0.50} $ & $ \boldsymbol{37.65 \pm 0.52}  $ & $ \boldsymbol{42.33 \pm 0.57} $ \\
        & \textbf{FedNoiL(Logarithm)}&  $ 41.51 \pm 0.52 $ & $ 47.12 \pm 0.47  $ & $ 41.46 \pm 0.55  $ & $ \boldsymbol{44.66 \pm 0.70}$ &  $ 34.12 \pm 0.80$ & $ 45.97 \pm 0.72 $ & $  37.21 \pm 0.46 $ & $ 41.68 \pm 0.60 $ \\
        \bottomrule
        \end{tabular}
        }
        }
    \end{table}
    \end{minipage}
    \vspace{-0.2in}
\end{figure}
(3) The dash lines in Fig.~\ref{fig:equal} (a) and (b) represent the constant schedules. The two decaying schedules of local epochs can effectively improve both the communication and computation-efficiency. (4) In Fig.~\ref{fig:parallel_1}, we compare the communication and computation efficiency trade-off between Cosine(80), Logarithm(80), and Constant(80). \textbf{Logarithm(80) achieves the same accuracy with the least communication rounds and training batches}. (5) In Fig.~\ref{fig:parallel_2}, we compare the trade-off between Logarithm schedules with $r_{\min}\in \{20, 40, 80\}$. At the end of training, schedules with a higher $r_{\min}$ are more efficient on both communication and computation. (6) In Fig.~\ref{fig:parallel_3}, we compare the trade-off between Cosine schedules with $r_{\min} = \{20, 40, 80\}$. A schedule with a larger $r_{\min}$ has higher communication efficiency but lower computational efficiency for achieving the same accuracy.

\subsection{Main Results}
According to Section~\ref{choice_schedule},
if we set $r_{\min} = 80$, FedNoiL with Cosine and Logarithm schedules have the highest communication and computation efficiency at the final round. Thus, we report the results of FedNoiL with these two schedules, denoted as \textbf{FedNoiL(Cosine)} and \textbf{FedNoiL(Logarithm)}, using $r_{\min} = 80$ for all the remaining experiments unless explicitly specified. Table~\ref{table:iid_main} shows the test accuracy of all approaches in all datasets with IID and non-IID distributions. We regard the method as converged if the accuracy difference between two consecutive rounds is smaller than 2\% over the last five rounds, and use the symbol $\ast$ to denote the results of the method that do not converge. FedNoiL(Cosine) and FedNoiL(Logarithm) outperform other methods by large margins in all settings. In the most challenging noise setting, Pair High noise in CIFAR100, only our method is stable. The convergence results and smaller standard deviations highlight the advantage of our method on robustness. \textbf{To compare NLL methods in the FL+NLL baselines}, those using DMix have higher accuracy and are more stable than those using JoSRC in most cases, indicating that DMix works better in combating noisy labels than JoSRC on local devices with insufficient clean samples. 
\textbf{To compare FL methods in the FL+NLL baselines}, 
those using FL counterparts other than FedAvg do not outperform those using FedAvg in all cases, implying that only improving the FL algorithm is not effective in improving the robustness of the global model in the presence of noisy labels. This is because personalized FL methods (e.g., APFL) are not designed for heterogeneous noise setting and the robust global aggregation methods (e.g., Krum and CFL) can not effectively prevent the global model from degradation with low-quality local models overfitting to the label noise.\looseness-1

\subsection{Extended Comparison in More Challenging Settings}
\begin{wrapfigure}{R}{0.6\textwidth}
\begin{minipage}{0.6\textwidth}
    \begin{table}[H]
        \centering
        \vspace{-30pt}
        \caption{\textbf{Test accuracy} (\%) of FedNoiL and baselines in settings with more clients and imbalanced samples in CIFAR-10.}
        \label{table:more_setting} 
        \vspace{0pt}
        
        \resizebox{1.0\textwidth}{!}{
        \setlength{\tabcolsep}{0.7mm}{
        \begin{tabular}{l|cc|cc|cc|cc}
        \toprule
         \multirow{3}{*}{Method} & \multicolumn{4}{c|}{100 clients} & \multicolumn{4}{c}{Imbalanced samples of clients}  \\ \cline{2-5} \cline{6-9}
        
         & \multicolumn{2}{c|}{Symmetric} & \multicolumn{2}{c|}{Pair} & \multicolumn{2}{c|}{Symmetric} & \multicolumn{2}{c}{Pair} \\ \cline{2-5} \cline{6-9}
         & high & low & high & low & high & low & high & low  \\  \midrule

        Vanilla FedAvg & $ 53.66^\ast $ & $   71.90 $ & $ 33.01^\ast  $ & $ 53.26^\ast $ & $ 37.85^\ast $ & $  55.78^\ast $ & $ 34.95^\ast  $ & $ 47.97^\ast $ \\

        FedAvg $+$ DMix&  $66.64  $ & $  70.65 $ & $ 60.34  $ & $ 65.10 $ & $ 59.96 $ & $  62.84 $ & $ 44.30^\ast  $ & $ 61.12 $ \\
        FedAvg $+$ JoSRC &  $59.67^\ast  $ & $   72.46 $ & $  55.61^\ast $ & $ 63.70^\ast$ & $ 39.11^\ast $ & $ 58.26^\ast $ & $ 36.70^\ast  $ & $ 49.80^\ast $ \\
        APFL $+$ DMix &  $ 75.05 $ & $ 79.41  $ & $ 66.26  $ & $77.94 $ & $ 64.70 $ & $69.06  $ & $  53.17^\ast$ & $ 70.08 $ \\
        APFL $+$ JoSRC  &  $  47.10 $ & $60.09   $ & $ 47.86^\ast  $ & $ 51.37^\ast$ & $ 30.62^\ast $ & $45.70^\ast  $ & $ 29.08^\ast  $ & $ 40.89^\ast $ \\
        Krum $+$ DMix &  $ 67.46 $ & $ 69.62  $ & $  58.30 $ & $60.99  $ & $ 61.10 $ & $ 63.86 $ & $ 40.60^\ast  $ & $ 59.01 $ \\
        Krum $+$ JoSRC &  $ 52.57^\ast $ & $ 64.12^\ast $ & $ 45.04^\ast  $ & $56.70^\ast $ & $ 45.90^\ast $ & $ 50.73^\ast $ & $  42.45^\ast $ & $ 53.09^\ast  $ \\
        CFL $+$ DMix &  $ 63.47$ & $ 67.85  $ & $  59.12 $ & $64.05 $ & $ 55.67 $ & $ 60.60 $ & $ 45.71  $ & $ 50.31 $ \\
        CFL $+$ JoSRC &  $ 55.30^\ast $ & $  62.79 $ & $  56.90^\ast $ & $60.39^\ast $ & $ 52.07^\ast $ & $ 59.64 $ & $  41.60^\ast $ & $ 53.05^\ast $ \\
        \cmidrule{1-9}
        \textbf{\textbf{FedNoiL(Cosine)}}&  $84.77  $ & $  88.22   $ & $ 75.62   $ & $\textbf{87.25}  $ & $ 74.68 $ & $79.61  $ & $ \textbf{69.02}  $ & $ 72.88 $ \\
        \textbf{\textbf{FedNoiL(Logarithm)}}&  $\textbf{85.94}  $ & $  \textbf{89.08}   $ & $ \textbf{76.81}   $ & $86.33  $ & $ \textbf{75.11} $ & $\textbf{80.01}  $ & $ 68.49  $ & $ \textbf{73.31} $ \\
        \bottomrule
        \end{tabular}
        }
        }
    \end{table}
\end{minipage}
\vspace{-15pt}
\end{wrapfigure}
To show the practical nature of FedNoiL, we extend it to more settings (More details can be found in Appendix) by conducting more analysis on CIFAR-10: \textbf{More clients}. We increase the number of clients to 100 (each having 500 samples) to cover the whole CIFAR10 dataset and compare FedNoiL with various baselines in Table~\ref{table:more_setting}. FedNoiL constantly outperforms all the baselines by a large margin, showing the robustness of FedNoiL in a large scale. \textbf{Imbalanced samples of clients}. We sample $q \sim Dirichlet(\beta)$ ($\beta=20$) and assign a $q_{k}$ proportion of the total samples to client-$k$. FedNoiL stably achieves SOTA
results while most baselines do not converge. \textbf{No label noise}. The test accuracy(\%) for FedAvg/FedNoiL on clean samples is 84.48/86.12 on CIFAR10. FedNoiL performs better since it selects easy samples to learn at first as a more effective curriculum.

\subsection{Ablation Study}
\noindent \textbf{Effects of Two-Level Sampling}.
To verify the effectiveness of our two-level sampling, we compare: (i) \textbf{FedNoiL(Original)}, which uses the proposed two-level sampling; (ii) \textbf{FedNoiL(Uniform Client Sampling)}, which chooses the clients uniformly while conducts local-data sampling as described in Section~\ref{intra_client}; (iii) \textbf{FedNoiL(Uniform Local-Data Sampling)}, which samples clients as described in Section~\ref{inter-client} while chooses samples within each chosen client using uniform probabilities. All these three approaches use the Logarithm schedule for a fair comparison.
Table~\ref{table:twostep} compares the quality of client sampling and local-data sampling of these three methods in all datasets under high symmetric and pair flipping noises. The average noise refers to the average real noisy ratio of all the selected clients at each round. Recall and Precision measure the quality of local-data sampling in all selected clients. 
\textbf{For client sampling}, the average noise ratio of FedNoiL(Original) is lower than FedNoiL(Uniform Client Sampling) in both symmetric and pair flipping noise. This indicates that the client sampling can select higher-quality local models associated with lower noise ratios for aggregation, resulting in a more robust global model. FedNoiL(Original) also has higher precision and recall since the better global model can estimate more accurate sampling probabilities for sample selection at the client end. \textbf{For local-data sampling}, the label precision and recall of FedNoiL(Original) are much higher than FedNoiL(Uniform Local-Data Sampling), indicating that our local-data sampling contributes greatly to the sample selection quality in the local dataset. We note that the poor sample selection performance of FedNoiL(Uniform Local-Data Sampling) also deteriorates the quality of the client sampling in the harder pair flipping case.

\begin{table*}[!h]
\centering

\caption{\textbf{Ablation Study} of two-level sampling on three datasets: comparisons of test accuracy, average noise ratio, label precision, and label recall (\%) of FedNoiL(Uniform Client Sampling), FedNoiL(Uniform Local-Data Sampling), and FedNoiL(Original) in high-noise ratio settings (symmetric and pair flipping noise).}\label{table:twostep}
\vspace{-0.08in}

\resizebox{1\textwidth}{!}{
\setlength{\tabcolsep}{0.6mm}{
\begin{tabular}{l|l|cccc|cccc}
\toprule
\multirow{2}{*}{Datasets} & \multirow{2}{*}{Method} & \multicolumn{4}{c|}{Symmetric High} & \multicolumn{4}{c}{Pair High} \\ \cline{3-10} 
& & Accuracy ($\uparrow$)  & Noise Ratio ($\downarrow$)& Precision ($\uparrow$)& Recall ($\uparrow$)& Accuracy ($\uparrow$)& Noise Ratio ($\downarrow$)& Precision ($\uparrow$)& Recall ($\uparrow$)\\ 

\midrule

\multirow{3}{*}{\begin{tabular}[c]{@{}c@{}}FASHION\\MNIST\\ \end{tabular}} 
 & FedNoiL(Uniform Client Sampling) & $ 81.71 \pm 0.45 $ & $ 65.78 \pm 3.60$ & $ 82.55 \pm 1.96 $ & $85.08 \pm 1.83 $ & $ 80.22 \pm 1.93 $ & $54.03 \pm 5.27 $ & $ 84.08 \pm 3.23 $ & $83.56 \pm 2.65 $  \\
 & FedNoiL(Uniform Local-Data Sampling) & $ 73.86 \pm 1.14 $ & $ 62.70 \pm 3.51 $ & $ 39.08 \pm 3.17 $ & $41.27 \pm 4.41 $ & $ 40.29^\ast $ & $58.25 \pm 6.65 $ & $ 35.86 \pm 6.61 $ & $ 40.90 \pm 3.99$  \\
 & FedNoiL(Original) & $\boldsymbol{82.46 \pm 0.54}  $ & $ \boldsymbol{62.29 \pm 2.82}$ & $ \boldsymbol{84.74 \pm 1.37}$ & $ \boldsymbol{88.05 \pm 1.61} $ &  $\boldsymbol{84.04 \pm 0.45} $ & $ \boldsymbol{47.79 \pm 6.00} $ & $ \boldsymbol{89.77 \pm 1.58} $ & $ \boldsymbol{88.87 \pm 1.54}$  \\
\midrule

\multirow{3}{*}{CIFAR-10}
 & FedNoiL(Uniform Client Sampling) & $ 73.42 \pm 0.70 $ & $65.88 \pm 3.56 $ & $ 73.19 \pm 2.67 $ & $75.02 \pm 1.98 $ & $60.58 \pm 3.58 $ & $ 55.26 \pm 5.89 $ & $63.76 \pm 5.29 $ & $ 63.49 \pm 3.94$  \\
 & FedNoiL(Uniform Local-Data Sampling) & $ 49.58 \pm 1.34 $ & $ 61.80 \pm 3.41 $ & $ 42.68 \pm 2.29 $ & $ 52.99 \pm 6.08 $ & $  48.41^\ast $ & $ 56.79 \pm 6.58$ & $39.21 \pm 6.09 $ & $ 41.25 \pm 2.16$  \\
 & FedNoiL(Original) & $ \boldsymbol{74.62 \pm 0.69} $ & $ \boldsymbol{59.17 \pm 3.52} $ & $ \boldsymbol{76.22 \pm 1.80} $ & $\boldsymbol{79.64 \pm 1.85} $ & $ \boldsymbol{71.44 \pm 0.68} $ & $ \boldsymbol{48.13 \pm 5.79} $ & $ \boldsymbol{77.72 \pm 2.69}  $ & $\boldsymbol{76.06 \pm 2.06 }$  \\
\midrule
\multirow{3}{*}{CIFAR-100}
 & FedNoiL(Uniform Client Sampling) & $40.56 \pm 0.28  $ & $64.98 \pm 2.79 $ & $ 69.27 \pm 2.67 $ & $ 72.93 \pm 5.11$ & $ 36.71 \pm 3.43 $ & $ 54.32 \pm 5.32$ & $  62.16 \pm 5.42 $ & $61.79 \pm 5.48 $  \\
 & FedNoiL(Uniform Local-Data Sampling) & $ 29.32 \pm 0.71 $ & $58.38 \pm 3.16 $ & $ 46.90 \pm 2.65 $ & $57.52 \pm 6.17 $ & $ 28.48^\ast $ & $56.00 \pm 6.89 $ & $ 42.55 \pm 8.41 $ & $ 47.57 \pm 5.63$  \\
 & FedNoiL(Original) & $ \boldsymbol{41.51 \pm 0.52} $ & $ \boldsymbol{55.79 \pm 2.48}$ & $\boldsymbol{75.29 \pm 2.67}  $ & $ \boldsymbol{79.30 \pm 3.42}$ & $\boldsymbol{41.46 \pm 0.55}  $ & $ \boldsymbol{47.30 \pm 5.78}$ & $\boldsymbol{71.36 \pm 4.47}  $ & $\boldsymbol{73.87 \pm 5.13} $  \\
\bottomrule
\end{tabular}
}
}
\vspace{-0.2in}

\end{table*}

\begin{figure*}[!h]
    \small
        \subfigure[Test Accuracy]{\includegraphics[width=0.24\textwidth]{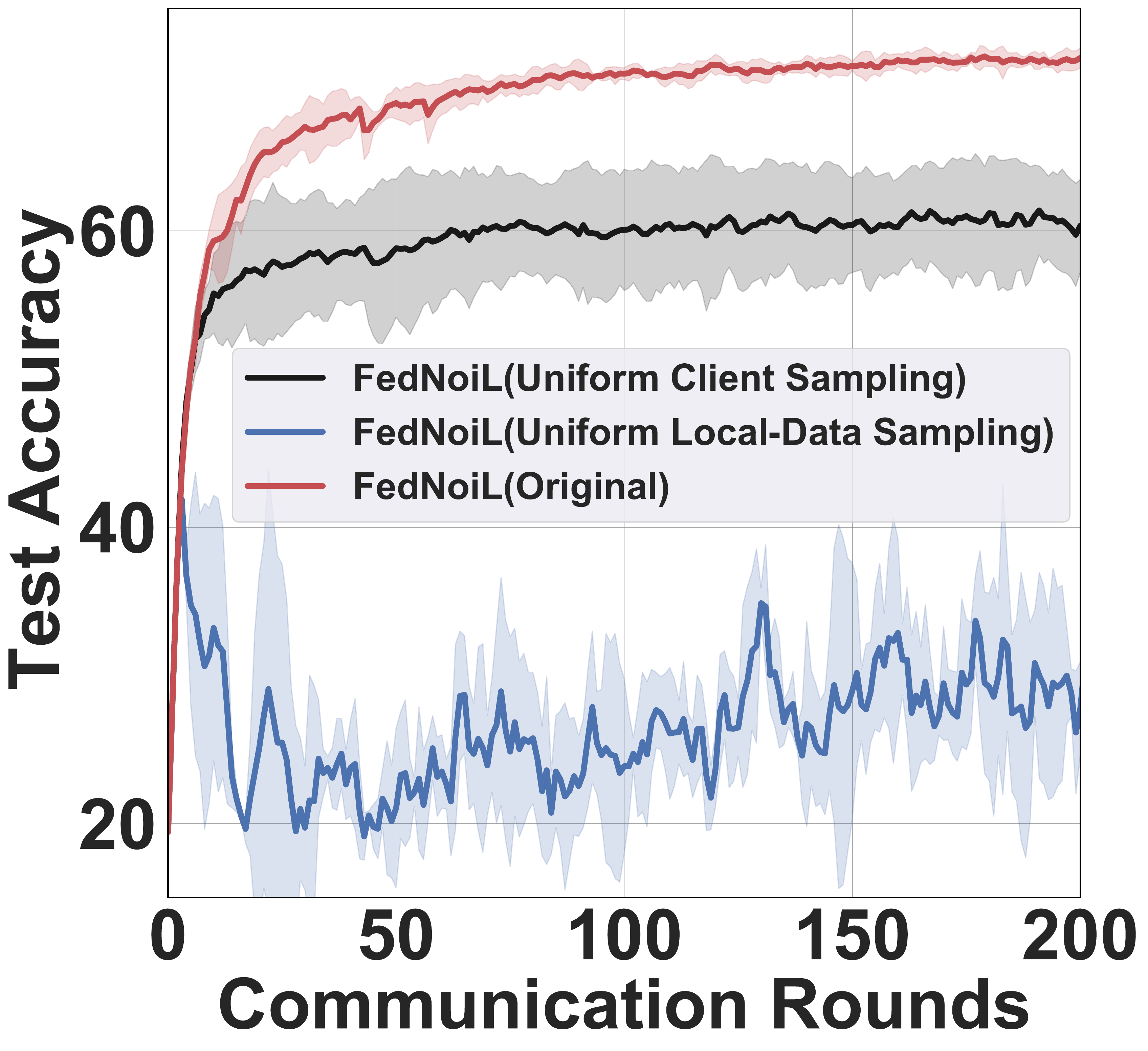}}
        \subfigure[Average Noise Ratio]{\includegraphics[width=0.24\textwidth]{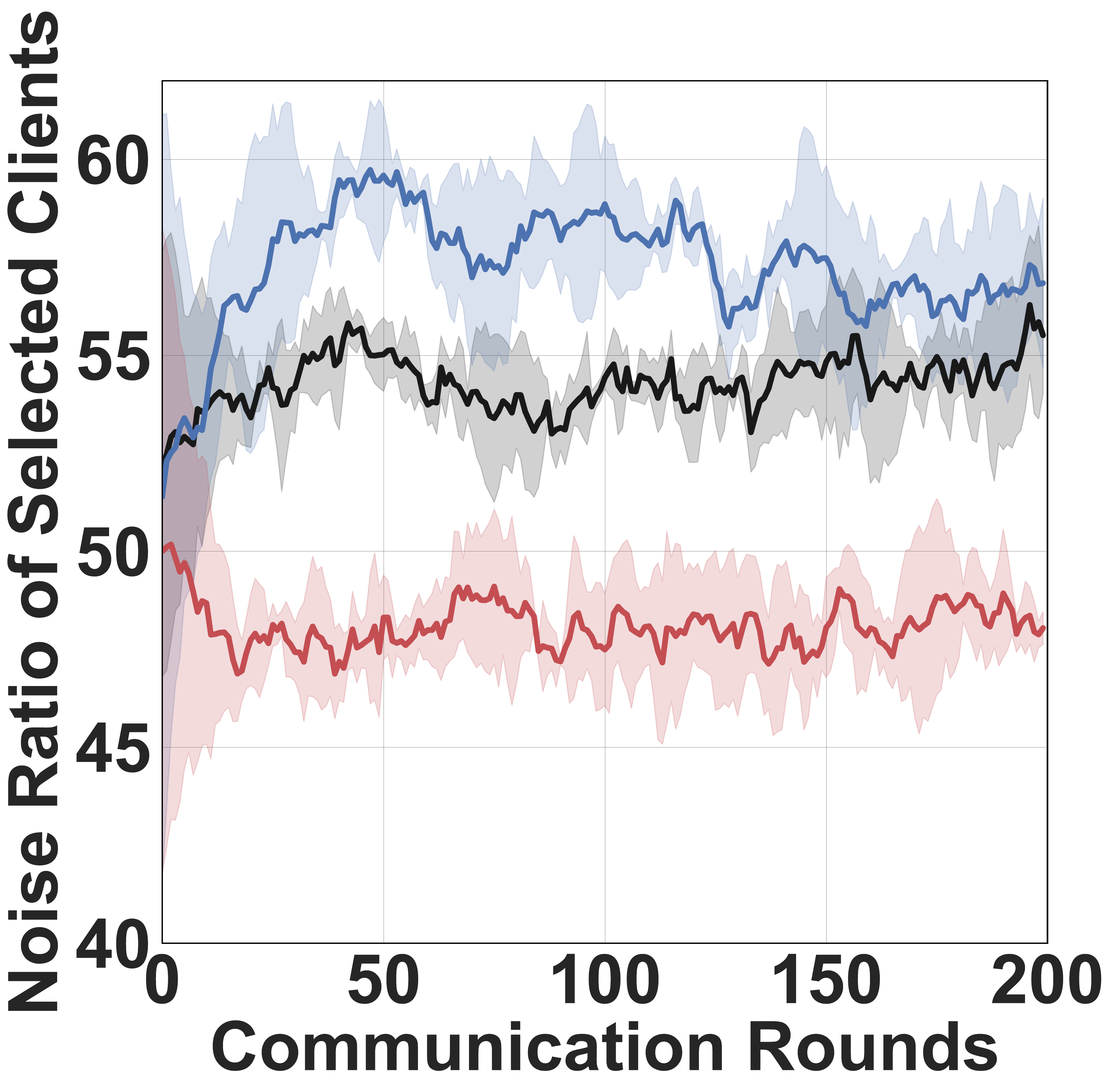}}
        \subfigure[Label Precision]{\includegraphics[width=0.24\textwidth]{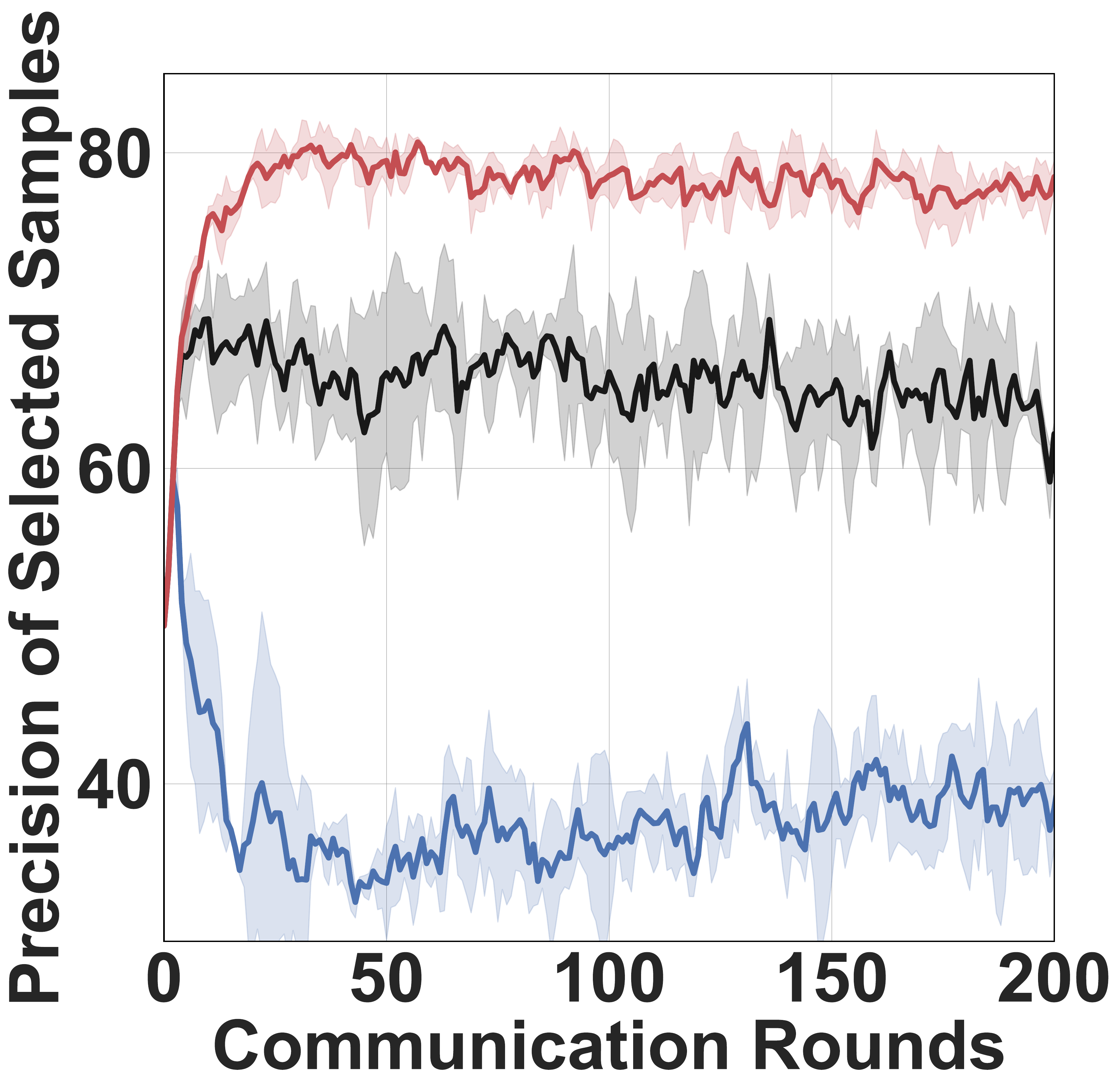}}
        \subfigure[Label Recall]{\includegraphics[width=0.24\textwidth]{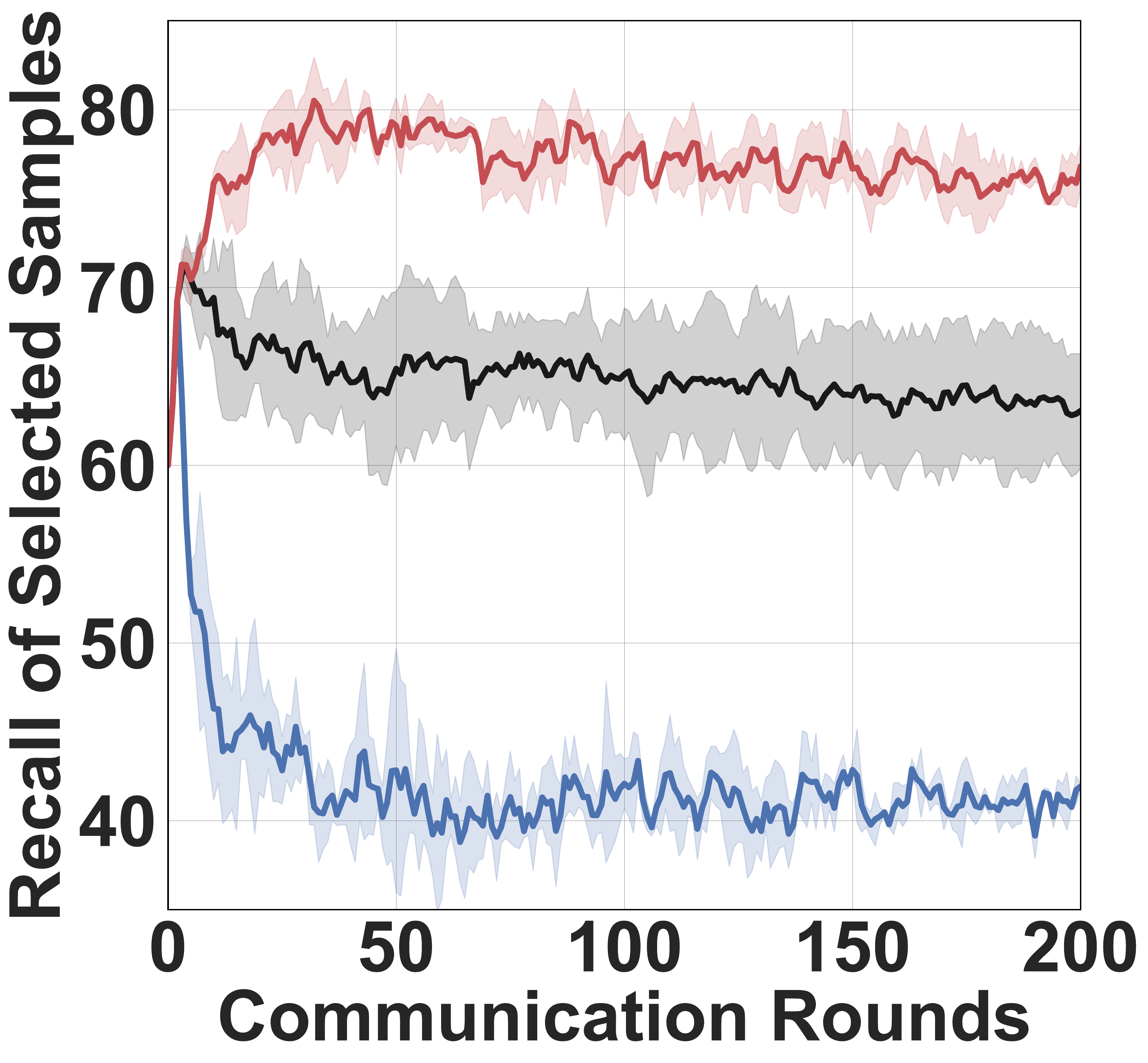}}
        \\
        
    \vspace{-0.15in}
    \caption{\textbf{Ablation Study} of two-level sampling on CIFAR-10: comparing the four metrics (in Appendix) of FedNoiL(Uniform Client Sampling), FedNoiL(Uniform Local-Data Sampling), and FedNoiL(Original) of high-noise (pair flipping) ratio setting.}
    \vspace{-0.1in}

    \label{fig:twostep}
\end{figure*}
Fig.~\ref{fig:twostep} shows the performance and sampling quality of the two-level sampling of these three approaches in CIFAR-10 at each communication round. FedNoiL(Original) outperforms the other two methods both in client sampling (lower average noise ratio) and local-data sampling (higher precision and recall) during the whole training process, demonstrating the benefits of the two-level sampling.
\begin{wrapfigure}{R}{0.5\textwidth}
\begin{minipage}{0.5\textwidth}
  \begin{table}[H]
        \centering
        \vspace{-30pt}
        \caption{Comparisons between FedNoiL\dag (without SSL) and FedNoiL on test accuracy (\%), label precision (\%), and label recall (\%) in CIFAR-10.}
        \label{table:fixmatch}
        \vspace{0pt}
        
        \resizebox{1.0\textwidth}{!}{
        \setlength{\tabcolsep}{0.7mm}{
        \begin{tabular}{l|l|cc|cc}
        \toprule
        \multirow{2}{*}{Metrics} & \multirow{2}{*}{Method} & \multicolumn{2}{c|}{Symmetric} & \multicolumn{2}{c}{Pair} \\ \cline{3-6} 
        & & high & low & high & low  \\ 
        \midrule
        \multirow{2}{*}{\begin{tabular}[c]{@{}c@{}}Test\\Accuracy\\ \end{tabular}}
         & FedNoiL\dag & $ 63.44 \pm 1.09 $ & $ 71.03 \pm 0.78$ & $ 62.14 \pm 1.49 $ & $ 67.39 \pm 0.68 $  \\
         & FedNoiL & $ \boldsymbol{74.62 \pm 0.69} $ & $ \boldsymbol{80.38 \pm 0.67} $ & $ \boldsymbol{71.44 \pm 0.68} $ & $\boldsymbol{75.66 \pm 0.76} $  \\
        \midrule
        \multirow{2}{*}{\begin{tabular}[c]{@{}c@{}}Label\\Precision\\ \end{tabular}}
         & FedNoiL\dag& $ 68.39 \pm 2.25 $ & $ 80.00 \pm 1.69 $ & $72.05 \pm 3.52 $ & $77.12 \pm 1.89 $  \\
         & FedNoiL & $ \boldsymbol{76.22 \pm 1.80} $ & $ \boldsymbol{84.36 \pm 1.52}$ & $ \boldsymbol{77.72 \pm 2.69} $ & $\boldsymbol{82.01 \pm 1.77}$  \\
        \midrule
        \multirow{2}{*}{\begin{tabular}[c]{@{}c@{}}Label\\Recall\\ \end{tabular}}
         & FedNoiL\dag & $ 71.08 \pm 2.37 $ & $ 82.78 \pm 1.63 $ & $ 70.18 \pm 2.60 $ & $76.52 \pm 1.85 $  \\
         & FedNoiL & $ \boldsymbol{79.60 \pm 1.85} $ & $\boldsymbol{87.04 \pm 1.61} $ & $ \boldsymbol{76.06 \pm 2.06} $ & $\boldsymbol{81.51 \pm 2.18}$  \\
        \bottomrule
        \end{tabular}
        }
        }
    \end{table}
\end{minipage}
\vspace{-15pt}
\end{wrapfigure}

\noindent \textbf{Additional Improvements by SSL}
To study the additional improvements brought by SSL in our method, we conduct an analysis that reveals the difference in performance when we remove SSL (denoted by FedNoiL\dag) at the client end. Table~\ref{table:fixmatch} shows the performance and sampling quality of our method with/without SSL in CIFAR-10. We use the Logarithm schedule for a fair comparison. Significant degradation in performance of FedNoiL w/o SSL is shown in all noise cases, especially for high noise case where more unselected samples can contribute to the training of local models. The benefits of SSL in our method are in these aspects: (i) leveraging the pseudo-labels of the unselected samples to improve the performance of the local model and (ii) improving the performance of clean sample selection incrementally.

\section{Conclusions}
In this paper, we study federated learning with local data corrupted by noisy labels, which is a practical but open challenge underexplored in literature. We propose a simple yet effective method, ``FedNoiL'', built upon a novel two-level sampling strategy, where the server samples high-quality local models for global aggregate and each client samples clean labels for local training. FedNoiL significantly outperforms simple combinations of SOTA FL and NLL methods on several benchmark datasets corrupted by different types of label noises in both IID and non-IID FL settings where in addition the noise ratios can vary drastically across clients.

\bibliographystyle{plain}
\bibliography{preprint}



\clearpage
\appendix

\end{document}